\documentclass[twoside,11pt]{article}

\usepackage[preprint]{jmlr2e}

\usepackage[utf8]{inputenc}
\usepackage{amsmath}
\usepackage{xcolor}
\usepackage{booktabs, multirow, multicol}
\usepackage{caption}
\usepackage{algorithm}
\usepackage{algpseudocode}
\usepackage{listings}
\usepackage{lastpage}
\expandafter\def\expandafter\UrlBreaks\expandafter{\UrlBreaks\do\/\do\*\do\-\do\~\do\'\do\"\do\-}
\usepackage{placeins}
\hypersetup{pdfborder={0 0 0}}

\lstdefinelanguage{json}{
basicstyle=\ttfamily\scriptsize,
numbers=left,
numberstyle=\tiny\color{gray},
stepnumber=1,
numbersep=8pt,
showstringspaces=false,
breaklines=true,
breakatwhitespace=true,
frame=single,
framesep=4pt,
rulecolor=\color{black!20},
string=[s]{"}{"},
stringstyle=\color{olive},
comment=[l]{//},
commentstyle=\color{black!50}\itshape,
}

\jmlrheading{}{2026}{1-\pageref{LastPage}}{}{}{}{Burak Susam, Ruiyuan Kang and Tingting Mu}

\ShortHeadings{AutoManifold: Agentic Design of Data Visualisation Algorithms}{Susam, Kang and Mu}
\firstpageno{1}

\begin{document}

\title{AutoManifold: Agentic Design of Data Visualisation\\ Algorithms via Manifold Embedding}

\author{\name Burak Susam \email burak.susam@postgrad.manchester.ac.uk \\
       \addr Department of Computer Science\\
       University of Manchester\\
       Manchester, M13 9PT, UK
       \AND
       \name Ruiyuan Kang \email kang@x32studio.com \\
       \addr X32 Studio\\
       Abu Dhabi, UAE
       \AND
       \name Tingting Mu \email tingting.mu@manchester.ac.uk \\
       \addr Department of Computer Science\\
       University of Manchester\\
       Manchester, M13 9PT, UK}

\maketitle

\begin{abstract}%
Manifold embedding for data visualisation is dominated by a small set of hand-designed state-of-the-art (SOTA) algorithms.
Their design targets are fixed at development time,  resulting in inflexible algorithmic behaviour that cannot be readily steered towards user-specified structural priorities. 
Hand-designing  new embedding algorithms to address user  priority requires highly specialised expertise and is time consuming.
To address autonomous algorithm design tailored to  user preference, we introduce an agentic algorithm generation pipeline AutoManifold.
It  composes  new manifold embedding algorithms from a constrained vocabulary of affinity, cost, and optimisation primitives, supported by multi-agent large-language-model (LLM) orchestration.
AutoManifold  conditions every stage of its design on user-specified structural preservation preferences, and iteratively refines the algorithm configuration through an LLM-guided, metric-grounded iterative loop.
We compare the generated algorithms  against the three strongest and most frequently used SOTA methods (t-SNE, UMAP, and PaCMAP)  on real-world datasets spanning different difficulty regimes  under identical evaluation infrastructure.
The results demonstrate that LLM-driven, priority-conditioned
algorithm synthesis can move beyond hyperparameter tuning, producing genuinely new  algorithms  that outperform traditional hand-designed methods in their ability to steer towards user-specified properties without compromising the intrinsic structure in the original data.
The code of AutoManifold is available at \url{https://github.com/buraksusam/AutoManifold_NeurIPS2026}.
\end{abstract}


\section{Introduction}

Manifold embedding is  widely used   in modern data analysis, underpinning exploratory visualisation,   e.g., in single-cell genomics, neuroscience, and the inspection of learned representations in deep neural networks. 
However, the research has shifted from a proliferation of new algorithms towards convergence on a small set of hand-designed methods.
Nowadays, the field is dominated  principally by t-SNE
\citep{vandermaaten2008}, UMAP \citep{mcinnes2018}, and PaCMAP
\citep{wang2021},  whose algorithmic targets and mechanisms are fixed at design time \citep{kobak2019art, becht2019umap, huang2022towards}. 
Each of these methods encodes a specific, immutable trade-off between local neighbourhood fidelity and global structural properties such as pairwise distance preservation and cluster separation \citep{wang2021, mu2017data}.
Such trade-off directly impacts the usefulness of the computed embeddings in  downstream data visualisation.
The practitioner's only lever for steering this
trade-off is through hyperparameter tuning over a narrow, algorithm-specific
parameter space. 
An expert-driven process of hand-designing a new embedding algorithm to better handle  such trade-off  is time consuming, due to the large design space and the difficulty of anticipating  effects of  new designs without  extensive trial-and-error experimentation.

Recent work has begun to explore whether large language models (LLMs) can act
as autonomous designers of numerical algorithms.
However, naive use of  LLMs as  code generators is     insufficient.
Effective algorithm design requires a more structured agentic pipeline.
Despite   promising results in
program synthesis, scientific code generation, and automated machine
learning \citep{romera2024,chen2023}, LLM-emitted algorithms are
prone to numerical pathologies, miscalibrated objectives, and silent
correctness failures.
Moreover, there is no systematic mechanism to ensure that the
generated methods can faithfully reflect  the user-specified design priorities rather than LLM training bias.

We explore the use  of LLMs to design new manifold embedding algorithms for data visualisation, which can  improve upon user-specified quality metrics. Our contributions include the following:
\textbf{(1)} We introduce a unified  embedding framework that identifies   core computational components governing embedding quality and provides structured formulation templates  for  data- and  metric-aware adaptation.
\textbf{(2)} We propose AutoManifold, an LLM-driven multi-agent  system that uses the embedding framework as algorithm skeleton, and enables  effective, reliable,  and controllable  algorithm generation with compositional  novelty.
AutoManifold is capable of handling nested tasks of novel function design, model configuration search, and hyperparameter optimisation, through creative and analytical reasoning capacity of LLMs. 
\textbf{(3)} We conduct comprehensive experiments to compare  AutoManifold-generated algorithms   with strong hand-designed baselines, evaluated from multiple complementary   perspectives through global, local, and outlier lenses.  
Results show that AutoManifold without priority constraint can design algorithms with more balanced and stronger performance than existing baselines, while priority-driven AutoManifold can  steer   synthesised algorithms towards  target metrics with the other properties remaining within an acceptable range.

\section{Related Work}
\label{sec:related}

\textbf{Manifold Embedding for  Data Visualisation:}
The modern landscape of manifold embedding   for data visualisation is shaped by a small number of influential algorithms, e.g., t-SNE \citep{vandermaaten2008} by minimising  the mismatch between similarity distributions, UMAP \citep{mcinnes2018}  by aligning fuzzy topological structures, and  PaCMAP \citep{wang2021} by  explicitly separating mid-near
pairs and far pairs to improve  trade-off between local and global fidelity.
Earlier methods such as Laplacian Eigenmaps \citep{belkin2003}, Isomap
\citep{tenenbaum2000}, and LargeVis \citep{tang2016} introduced key designs, such as k-nearest-neighbour (kNN) graphs, geodesic distances, and 
negative-sampling gradients, which recur in  later algorithms.
Despite their differences, these methods hardcode  specific ways for   neighbourhood modelling,  objective formulation, and   optimisation, meaning  that  their inductive biases are fixed at design time. 

\textbf{LLM for  Data Visualisation:} Recent research has begun to explore the use of LLMs in data
visualisation, focusing on generating, interpreting, and interacting with
visual representations.
Systems like LIDA \citep{dibia2023}, MatPlotAgent
\citep{yang2024matplotagent}, and  PlotGen \citep{goswami2025} generate visualisations, while Charts-of-Thought \citep{chartsofthought2025} and LLM4Vis \citep{wang2023llm4vis} interpret and recommend  visualisations.
Survey studies \citep{nlp4vis2024, brossier2026} further investigate interactions between LLMs and visualisation.
However, in all such work, the embedding step that
computes the visualisation coordinates  is   treated as a black box, where an off-the-shelf algorithm   is called with default or lightly tuned parameters.
The role of LLMs  begins only after embedding computation.
Our earlier work \citep{susam2026explainable} uses an LLM agent to iteratively refine the hyperparameters of an existing embedding algorithm with explanations, but leaves the algorithm itself unchanged.

\textbf{LLM-Driven Algorithm Design:}
A rapidly growing body of work explores  LLMs
as autonomous designers of numerical algorithms  \citep{liu2024survey}. 
Existing frameworks span a spectrum of decreasing LLM autonomy, from unconstrained code generation to tightly scoped component optimisation.
At the high-autonomy end, the LLMs generate  entire algorithms from scratch, as in OPRO \citep{yang2023opro} for optimisation, LLaMEA
\citep{vanstein2024} for   metaheuristic design, and EvoTune \citep{surina2025} for enhancing evolutionary  search.
At the low-autonomy end, humans design fixed algorithmic structure, while   LLMs optimise specific  components within the structure.
For instance, FunSearch \citep{romera2024} evolves a  critical function by using LLMs within a provided algorithm skeleton for improving  cap-set construction and  bin-packing heuristics,  ReEvo \citep{ye2024} introduces  ``verbal gradients'' for improving combinatorial optimisation, and Eureka \citep{ma2023eureka} applies the same paradigm to  improve reward  design in reinforcement learning.
 LLMs have also made  progress in symbolic  discovery  \citep{shojaee2024} and automated feature engineering \citep{hollmann2023}.
However, most existing LLM-driven design frameworks optimise a  single  fixed 
 objective, with limited mechanism   to steer
the design  process towards  user-specified  metrics.

\section{A Unified Manifold Embedding Design Framework}
\label{sec:framework}

We focus on   computing low-dimensional embeddings $ \{\mathbf z_i\in \mathcal{R}^d \}_{i=1}^n$ for a  given set of high-dimensional data points $ \{\mathbf x_i\in \mathcal{R}^D\}_{i=1}^n$, where $d=2,3$ is used for data visualisation.
We seek low-dimensional  embeddings that preserve local neighbourhood relationships among the high-dimensional  data points, assuming an underlying manifold structure.
Embedding quality is examined by  reliability scores  denoted by $\textmd{score} \left(\left\{\left(\mathbf x_i, \mathbf z_i\right) \right\}_{i=1}^n \right)$. 
Encouraged by the rapid growth of LLMs with their increasingly powerful search and knowledge-driven reasoning capacity, we revisit manifold embedding algorithm design, and propose \textit{a unified, high-capacity embedding framework}, integrating a broad range of key factors and computational steps that influence the embedding quality to formulate a challenging optimisation problem.
This problem provides a testbed to agentic AI on   novel  function design, model configuration search, and hyperparameter optimisation.
This framework serves as an algorithmic skeleton for  algorithm synthesis, guiding the autonomous LLM designer.
Below we explain the proposed embedding framework component by component.

\textbf{Embedding Difficulty Analysis:} 
Significant dimensionality reduction  inevitably introduces information distortion when representing complex high-dimensional data patterns.
Local positioning of high-dimensional data points   strongly influences  the embedding difficulty, which can vary  across different regions of the data manifold.
Motivated by this, we introduce a regional  strategy that distinguishes  data patches with varying levels of embedding difficulty, and  adapts the model design accordingly.
Specifically, we assign data points into groups that are ``hard-to-embed" (H), ``moderately difficult to embed"(M), and ``easy-to-embed" (E),  and use the resulting difficulty labels as   inputs to guide the formulation of relevant framework components. 

Starting from clustering  $\left\{\mathbf x_i  \right\}_{i=1}^n$ into local patches (e.g., by k-means clustering),  we   embed each patch  through a standard embedding algorithm (e.g., t-SNE or UMAP) with its default hyperparameter setting.
Subsequently, we compute a set of reliability scores $\{\textmd{score}_l\}_{l=1}^L$ for each patch, and then assign each patch a difficulty level based on the  aggregated score. 
This process can be exemplified by  the function $\textmd{diff}(\mathbf x_i): \mathcal{R}^D\rightarrow \{\textmd{H}, \textmd{M}, \textmd{E}\}$, given as  
\begin{equation}
    \textmd{diff}(\mathbf x_i) = \textmd{assign}\left(\sum_{l=1}^L  \textmd{score}_l\left(\left\{\left(\mathbf x_j, \mathbf z^{\textmd{t-SNE, UMAP}}_j\right) \right\}_{\textmd{clustID}(\mathbf x_j )=\textmd{clustID}(\mathbf x_i)} \right) \right).
\end{equation}
Here, the difficulty assignment function is  to be designed, of which
an example design is a simple thresholding function, assigning ``H'' to low values of the aggregated score, while  ``E'' to high values and ``M'' to middle values.  
The cluster partition $\textmd{clustID}(\mathbf x )$, the standard embedding computation, and  the used   reliability scores  $\{\textmd{score}_l\}_{l=1}^L$   are implemented as fixed framework components.
Difficulty is assigned per patch rather than per point, since embedding difficulty arises from the relative distance relationships within a neighbourhood rather than from individual points.

\textbf{Affinity Matrix Construction:}  A common representation of a neighbourhood graph is the affinity matrix $\mathbf{A}=[a_{ij}]\in \mathcal{R}^{n\times n} $, where $a_{ij}$ measures the similarity between points $i$ and $j$, and  $a_{ij}=0$ indicates the absence of a neighbouring relationship.
Construction of such affinity matrix   usually depends on   choices of   similarity measures (e.g., Euclidean distance, cosine similarity  or geodesic distance), and  definition of  neighbourhood relationships based on these similarities (e.g.,  mutual or directed  kNNs, or probabilistic neighbourhoods).
Letting $Q$ denote the total number of affinity matrix configurations resulting from these choices,  each configuration is defined by  model choices stored in $\textmd{OptA}_l$  associated with fixed hyperparameters stored in $\theta^{(A)}_l$.
We construct an ensemble of affinity matrices computed under  different   configurations.
The aggregated affinity value between a target point $i$ and its potential neighbouring point $j$ is computed by  
\begin{equation}
\label{affine_high}
    a_{ij} = \sum_{l=1}^Q w_l\left(\mathbf x_i, \mathbf{f}^{(A)}(\mathbf x_i), \textmd{diff}(\mathbf x_i), \mathbf{g}\left (\{\mathbf{x}_i\}_{i=1}^n\right)\right) a_l\left(\mathbf x_i, \mathbf x_j, \textmd{OptA}_l, \theta^{(A)}_l \right).
\end{equation}
When each weight  $0\leq w_l\leq 1$ is a  constant that does not change over   data points, the affinity ensemble is simply a weighted sum of some base affinity matrices, referred to as \emph{hybrid affinity}.
More generally, we introduce \emph{adaptive weighting}, allowing each  weight to change  over   data points. 
Given a high-dimensional point $\mathbf x_i$, the  weight functions adapt to its embedding difficulty level $\textmd{diff}(\mathbf x_i)$ and its geometric and distributional features extracted by the feature mapping function $\mathbf{f}^{(A)}(\mathbf{x}):\mathcal{R}^D\rightarrow \mathcal{R}^{h_A}$.
They also take into account the overall data characteristics, represented by features describing the geometry, density and cluster imbalance issues of the input  data.
These features are extracted by a characteristic function $\mathbf{g}:\mathcal{R}^{n\times D}\rightarrow \mathcal{R}^h$, for which a pre-defined feature list is suggested.
The adaptive weight, feature mapping, and characteristic functions are to be designed. 
Each affinity function $a_l$ under a given configuration is implemented as fixed framework components, while the configuration options $ \{\textmd{OptA}_l\}_{l=1}^Q $ are  pre-defined, with their corresponding hyperparameters $ \{  \theta^{(A)}_l\}_{l=1}^Q $   to be determined.

\textbf{Embedding Optimisation Formulation:}
In general, distortion of neighbourhood relationships can be   quantified by comparing   two  affinity matrices, of which one is constructed in the original space (e.g., by Eq. \ref{affine_high})  while the other in  the embedding space (typically derived from Euclidean distances).
This  motivates constructing \emph{local objective functions} for   embedding optimisation (e.g., similarity distribution based  in t-SNE and fuzzy topological structure based   in UMAP).
To balance local fidelity with  global  structure preservation, additional \emph{global objective functions} are introduced to capture large-scale geometry or distributional properties  (e.g., variance-based  in  PCA, and   mutual information or statistical dependence based).

Given the multi-faceted nature of these objectives, we formulate   embedding optimisation   as a multi-objective problem, combining a set of $H$  objective functions, each of which is denoted at  point level by $O_l(\mathbf{x}_i, \mathbf{z}_i, \textmd{OptC}_l)$ for $l\in\{1,2\ldots, H\}$, with  pre-defined objective  configurations $\{\textmd{OptC}_l\}_{l=1}^H$.
This results in a standard multi-objective problem, as 
\begin{equation}
\label{embedding_opt_multi}
    \min_{\{\mathbf{z}_i\}_{i=1}^n} \left(\delta_{1}\sum_{i=1}^nO_1(\mathbf{x}_i, \mathbf{z}_i, \textmd{OptC}_1), \ldots, \delta_{H}\sum_{i=1}^nO_H(\mathbf{x}_i, \mathbf{z}_i, \textmd{OptC}_H)\right),
\end{equation}
where $\delta_{l} \in \{0,1\}$ is a binary selector, deciding the inclusion/exclusion of an objective function.
A solver usually seeks the Pareto front of the problem.
An alternative   formulation is to optimise the weighted sum of these objective functions, resulting in the following unconstrained   problem:
\begin{equation}
\label{embedding_opt_unconstrained}
    \min_{\{\mathbf{z}_i\}_{i=1}^n}  \sum_{l=1}^H\sum_{i=1}^n \lambda_l\left(\mathbf x_i, \mathbf{f}^{(O)}(\mathbf x_i), \textmd{diff}(\mathbf x_i), \mathbf{g}\left (\{\mathbf{x}_i\}_{i=1}^n\right), \{\textmd{score}_l\}_{l=1}^L \right)  O_l(\mathbf{x}_i, \mathbf{z}_i, \textmd{OptC}_l),
\end{equation}
where  we adopt the similar adaptive weighting scheme as in affinity construction, for combining  objective functions.
Each weight $0\leq \lambda_l\leq 1$ can either be a constant unchanging over data points, or a function depending on the embedding difficulty level and feature characteristics of a given point, as well as data characteristics of the whole input data  and the embedding quality metrics of interest.
We adopt the same characteristic function as used for affinity matrix construction, but use a different instance-level feature mapping function  $\mathbf{f}^{(O)}(\mathbf{x}_i):\mathcal{R}^D\rightarrow \mathcal{R}^{h_O}$.
For example,   a density feature  $f_{\text{density}}(\mathbf{x}_i)$ normalised between 0 and 1 can be extracted for each data point.
An adaptive weight function in the form of $\lambda_l = 1- f_{\text{density}}(\mathbf{x}_i)$   can then enforce  stronger local neighbourhood preservation for points from sparse regions.

Typically,   the primary objective of  manifold embedding is  to learn a low-dimensional representation that  preserves the local geometry around each data point.
Therefore, objective functions that emphasise local structure often play a dominant role in embedding calculation.
Global structure preservation is often incorporated to  maintain large-scale  geometry, acting as a complementary constraint or   regularisation term. 
This encourages us to introduce a constrained formulation,   emphasising  primary objective optimisation  while treating the complementary  objectives as optimisation  constraints. 
We distinguish the index sets of the primary and complementary objective functions by $I_p$ and $I_c$, and optimise the embeddings by 
\begin{align}
\label{embedding_opt_constrained}
    \min_{\{\mathbf{z}_i\}_{i=1}^n} \; \;& \sum_{l\in I_p} \sum_{i=1}^n \lambda_l\left(\mathbf x_i, \mathbf{f}^{(O)}(\mathbf x_i), \textmd{diff}(\mathbf x_i), \mathbf{g}\left (\{\mathbf{x}_i\}_{i=1}^n\right), \{\textmd{score}_l\}_{l=1}^L\right)  O_l(\mathbf{x}_i, \mathbf{z}_i, \textmd{OptC}_l), \\
    \nonumber
    \text{s.t. } & \sum_{i=1}^n    O_l(\mathbf{x}_i, \mathbf{z}_i, \textmd{OptC}_l) \leq \theta_l, \textmd{for } l\in I_c. 
\end{align}
We can also formulate a multi-objective  problem with the same constraints  as an alternative to the   above weighted sum.
By introducing constraints, we do not  enforce the  minimisation of complementary  objective functions, but restrict  their values to be sufficiently small controlled by hyperparameters $\{\theta_l\}_{l=1}^H$.
The choice of which optimisation formulations in Eqs. (\ref{embedding_opt_multi}) to (\ref{embedding_opt_constrained}) to use and which objective functions to consider indicated by $\delta_l$ for Eq. (\ref{embedding_opt_multi}) are to be searched.
Both the weights $\{\lambda_l\}_{l=1}^H$ and  feature mapping function $\mathbf{f}^{(O)}$  are to be designed.  
Although the formulation logic of  each individual embedding objective function $O_l$ is specified by its configuration, its exact function form still needs to be calibrated accordingly.

\textbf{Model Configuration Determination:} To construct the affinity matrix and optimisation formulation, configuration choices are searched from   the candidate pools $\{\textmd{OptA}_l\}_{l=1}^Q$  and $\{\textmd{OptC}_l\}_{l=1}^H$.
We introduce a two-stage decision making, where the final configuration choice depends on an initial configuration guess informed by input data characteristics.
This results in 
\begin{align}
  \label{eq:optA}
    \textmd{OptA} =\; & F_{\textmd{OptA}}\left(\textmd{OptA}^{(0)}, \{\textmd{OptA}_l\}_{l=1}^Q, \{\mathbf{x}_i\}_{i=1}^n\right), \\
    \label{eq:optC}
    \textmd{OptC} = \;& F_{\textmd{OptC}}\left(\textmd{OptC}^{(0)}, \{\textmd{OptC}_l\}_{l=1}^H, \{\mathbf{x}_i\}_{i=1}^n, \{\textmd{score}_l\}_{l=1}^L\right).
\end{align}
We let Eq. (\ref{eq:optC}) take   extra input $\{\textmd{score}_l\}_{l=1}^L$, allowing prior knowledge on embedding quality metrics to influence  the objective configuration  for embedding optimisation.
The initial  choices of  affinity   and objective configurations are informed by data characteristics of the high-dimensional data represented by  $\mathbf{g}$, such as
\begin{equation}
    \left(\textmd{OptA}^{(0)}, \textmd{OptC}^{(0)}\right) = F_{\text{O-ini}}(\mathbf{g}\left (\{\mathbf{x}_i\}_{i=1}^n)\right).
\end{equation}
The configuration functions $F_{\textmd{OptA}}$ and  $F_{\textmd{OptC}}$   and the initialisation function $F_{\text{O-ini}}$  are to be designed.

\textbf{Embedding and Hyperparameter Optimisation:} Regarding to solving the three numerical optimisation problems in Eqs. (\ref{embedding_opt_multi}) to (\ref{embedding_opt_constrained}), well-established solvers (e.g., gradient-based methods and their accelerated versions, pareto-based methods) are available. 
We provide a pre-defined list of solvers for the framework to choose.
Regarding  hyperparameter optimisation,  we introduce an \textit{iterative refinement} process,  building on the LLM-agent refinement architecture of our earlier work \citep{susam2026explainable}. Starting from an initial hyperparameter guess  $  \theta^{(0)}$,  it iteratively updates  by
\begin{equation}
\label{eq:iterative_refine}
      \theta^{(t+1)}  = F_{\text{update}} \left(   \theta^{(t)},  \{\textmd{score}_l\left(\{(\mathbf{x}_i,\mathbf{z}_i )\}_{i=1}^n\right)\}_{l=1}^L     \right).
\end{equation}
Here, the updating function $F_{\text{update}}$ refines the current  hyperparameters by considering reliability scores of the current embedding results.
The updating function is to be designed.

\section{Multi-Agent Algorithm Synthesis  }
\label{sec:method}

The  proposed manifold embedding framework in Section \ref{sec:framework} has high modelling capacity and adapts to the data, but it introduces a large search space of model configurations and  hyperparameters.
Moreover, it introduces novel design tasks of specifying     functions, which may be  given as explicit analytical formulas, described by explicit heuristic rules, or implemented as implicit black-box models.
These include design cases of difficulty assignment function (explicit), weight functions $w_l$ and $\lambda_l$ (explicit), feature mapping functions $\mathbf{f}^{(A)}$, $\mathbf{f}^{(O)}$ and $\mathbf{g}$ (explicit),  also configuration, initialisation, and hyperparameter  updating  functions $F_{\textmd{OptA}}$, $F_{\textmd{OptC}}$, $F_{\text{O-ini}}$  and $F_{\text{update}}$   (implicit), as well as the calibration of $O_l$ (explicit).
Such a complex combination of nested search and design tasks poses a significant challenge for human algorithm developers to complete within a short time frame.
To address this challenge, we present AutoManifold, a principled algorithm synthesis pipeline,  driven by an LLM based  multi-agent  system.
We delegate creative and analytical reasoning  to LLM agents while measuring quantitative performance objectively by pre-implemented, deterministic, non-LLM engines. 
We strictly follow the central design principle of \textit{``agents propose, engines measure''}.
AutoManifold allows explicit user control on which embedding metrics for the algorithm to improve over.
Specifically, the user is asked to select one of the eight pre-defined priority options as listed in Table \ref{tab:priority} of Appendix \ref{app:user}.
AutoManifold  generates executable code of the designed  algorithm, underpinned by   mathematical formulation of   embedding optimisation  based on templates provided in Eqs. (\ref{embedding_opt_multi}) to (\ref{embedding_opt_constrained}) and by structured pseudo-code at the granularity of algorithmic steps.
\textbf{Orchestrated Architecture:} AutoManifold is organised into three independent layers:
The \textit{orchestration layer} contains a  planner agent, which codes the phase-by-phase workflow logic, manages pipeline state, and enforces  convergence criteria for the iterative refinement.
The  \textit{agent layer} contains  specialised LLM-backed worker agents,
 each constrained to a narrow task scope through a structured system
  prompt and a required JSON output schema.
The  \textit{compute engine layer} contains   pre-implemented deterministic engines,    computing embeddings and  reliability scores for a given embedding algorithm  and  a dataset.
These engines serve as objective validators for every LLM-generated output.
Each worker is a single-shot, schema-constrained LLM call that performs no embedding, metric computation or tool call of its own.
The planner invokes the engines and injects their outputs verbatim into the context of the next agent, so that agents cannot fabricate metrics.
The phase graph is fixed by design, as a reliability choice over a dynamically orchestrated controller.
More details on our adopted agentic workflow are provided in Appendix~\ref{app:workflow}.
The system operates through a pre-analysis phase, followed by six processing phases including (1) baseline analysis, (2) data characterisation, (3) data structure analysis, (4) algorithm design, (5) algorithm synthesis, and (6) iterative hyperparameter refinement, post analysis and output, which are implemented  by the planner agent with the assistance of  multiple worker agents.
\begin{figure}[t]
  \centering
  \includegraphics[width=\textwidth]{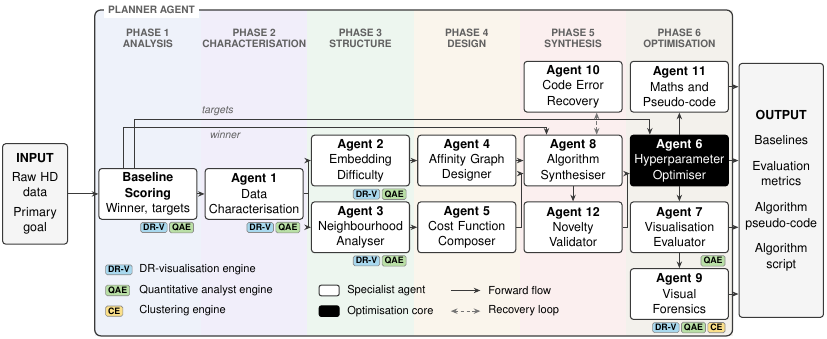}
  \caption{Overview of the multi-agent LLM pipeline for manifold embedding algorithm design.}
  \label{fig:pipeline}
\end{figure}

\textbf{Agent Role Distribution:} Each agent independently decides its goals or actions, with its decision making   informed by other agents when needed. 
The planner agent takes as input the user-specified embedding quality metrics,  the JSON output returned by the worker agents,  implements actions by generating code and calling compute engines, and runs the code to calculate   embeddings and evaluate their performance.
In \textbf{phase 1}, the DR-visualisation and quantitative analyst engines run baseline embedding methods (e.g., PCA, t-SNE, UMAP, etc.) on the data, to determine the best baseline and the targets of the optimisation loop.
The scores of the best baseline are passed to the algorithm synthesiser.
In \textbf{phase 2},  \textit{Agent 1}  designs the data characteristic function $\mathbf{g}$,   the implicit configuration initialisation function $F_{\text{O-ini}}$ based on  $\mathbf{g}$, and recommends a data pre-processing option (e.g., by PCA to reduce dimensionality to $d<m<D$).
In \textbf{phase 3},   \textit{Agent 2}  designs the difficulty assignment function $\textmd{diff}(\mathbf x)$, and \textit{Agent 3}  performs neighbourhood analysis on the pre-processed data.
Phases 2 and 3  together constitute a pre-processing and pre-analysis stage.
\textbf{Phase 4} conducts the main design task, supported  by \textit{Agent 4} for designing the affinity matrix based on Eq. (\ref{affine_high}), \textit{Agent 5} for designing  the optimisation formulation based on Eqs. (\ref{embedding_opt_multi}) to (\ref{embedding_opt_constrained}), and \textit{Agent Planner} for auto-selecting a solver for the composed optimisation problem.
We separate the options of determining constant weights from designing adaptive weight functions, to encourage the agents to pursue both routes.
As  a by-product result, Agents 4 and 5 implicitly design $F_{\textmd{OptA}}$ and $F_{\textmd{OptC}}$.
In \textbf{phase 5} for algorithm synthesis,  \textit{Agent 8} is the synthesiser, assisted by an independent \textit{Agent 10} for recovering  execution failures. 
\textit{Agent 12}  validates the novelty of the designed algorithm, determining the most similar method by comparing with the  mathematical
specifications of existing embedding methods (e.g., t-SNE, UMAP, PaCMAP, LargeVis \citep{tang2016}, Spectral Embedding \citep{belkin2003}, and ForceAtlas2 \citep{jacomy2014forceatlas2}) and analysing the fundamental differences.
The  next  phase  constitutes the algorithm refinement, post-analysis and output stage.
In \textbf{phase 6},   \textit{Agent 6} takes the synthesised embedding algorithm and  iteratively refines its hyperparameters following Eq. (\ref{eq:iterative_refine}), based on performance evaluation conducted by \textit{Agent 7}, who is provided with a list of  embedding reliability scores of interest.
Next, \textit{Agent 11}  prepares the mathematical formulation, executable code, and pseudo-code for the designed algorithm, and \textit{Agent 9} performs a post-hoc visual forensics.
This workflow is illustrated in Figure \ref{fig:pipeline}.

\textbf{Algorithm Novelty:}  AutoManifold generates genuinely novel manifold embedding algorithms through a composition of algorithmic  components guided by an algorithm skeleton, i.e., the unified embedding framework. 
 LLMs are provided with a fixed set of configuration options, referred to as a primitive vocabulary, while being free to combine
primitives in novel ways under a  set of suggested composition patterns.
Therefore, we consider the novelty   aimed by  AutoManifold as compositional novelty,  meaning new algorithms mostly built from novel combinations of known components.
It is worth to mention that, although the prompt files provide primitives covering design choices for  affinity formulations,  optimisation objectives,  constraints  and  solvers, these only serve as guidance rather than hard constraints.
In practice, LLM agents can depart from the provided choices and generate novel  formulations when they better optimise the target metrics, reflecting the  non-deterministic nature of LLM-based design.

\section{Experiments and Result Analysis}
\label{sec:experiments}

We compare the new  algorithms designed by AutoManifold against three strong SOTA manifold embedding algorithms, including t-SNE, UMAP  and PaCMAP with their hyperparameters tuned by random-search \citep{bergstra2012random} to maximise a balanced objective (a weighted sum of multiple scores) as detailed in Appendix~\ref{app:others}.
They are evaluated on six real-world datasets spanning different difficulty regimes, including one easy dataset (iris), three moderately difficult datasets (aging, kidney, pbmc3k), and two hard datasets (fashion-mnist, cortex). 
All agents use Gemini-3.1-Pro as the LLM backbone; alternative backbones are compared in Appendix~\ref{app:backbone_skeleton}.

\textbf{Embedding Evaluation:} We assess algorithm performance using six  reliability scores common for benchmarking embedding quality,  which   also    act  as $\{\textmd{score}_l\}_{l=1}^L$ in the embedding framework.
They are implemented as deterministic code  for evaluation, while provided as text input to   agents for driving algorithm design.
These include  Spearman  rank correlation  ($\rho$) and stress  ($S$) for assessing global  preservation of pairwise distances, silhouette score for measuring cluster separation ($C_{\text{sil}}$),    trustworthiness  ($T_{\text{area}}$)  and continuity  ($C_{\text{area}} $)  for examining local  fidelity over  varying neighbourhood sizes, and median local outlier factor (LOF) ($L$) for examining   outlier amount in   embedding space.
Four additional scores, which are never provided to the agents, are used in  held-out, stability and ablation experiments, including Shepard correlation (Shep) for distance preservation, normalised mutual information (NMI) for cluster agreement with ground-truth labels, RNX for local neighbour preservation, and kNN classification accuracy for  local neighbour consistency.
All scores are defined in Appendix~\ref{app:metrics}.

\textbf{Primitive Vocabulary:} 
We define primitive vocabulary to communicate model configuration options to   agents  in natural language.
These include  six  primitives $\{\text{OptA}_i\}_{i=1}^6$ for constructing  affinity matrix, five primitives $\{\text{OptC}_i\}_{i=1}^5$ for specifying   objective function,  and four primitives $\{\text{OptS}_i\}_{i=1}^4$ for optimisation solvers. 
Further details on datasets, difficulty definition, reliability scores, primitive vocabulary, and implementation  are provided  in the appendix, along with the complete prompt files for each agent.

\begin{table*}[t]
  \centering
  \caption{Priority-controlled algorithm synthesis. \textbf{Bold} indicates the best   across all   methods on that  dataset/score. \underline{Underline} flags all AutoManifold-generated algorithms (Auto-$\ast$) that outperform  the best  SOTA baseline  (t-SNE, UMAP or PaCMAP).$\uparrow$/$\downarrow$: higher/lower is better.
  }
  \label{tab:priority-steering}
  \setlength{\tabcolsep}{2pt}
  \resizebox{\textwidth}{!}{
  \begin{tabular}{llcccccccccccc}
    \toprule
    Dataset & Score & t-SNE & UMAP & PaCMAP & Auto  (w/o) & Auto-T & Auto-C & Auto-$\rho$ & Auto-Sil & Auto-S & Auto-L & Auto-local & Auto-global \\
    \midrule
    \multirow{6}{*}{\shortstack[l]{\texttt{iris}\\ \citep{Unwin2021TheID}}} & $T_{\text{area}}\uparrow$ & 0.9848 & 0.9810 & 0.9483 & \underline{0.9865} & 0.9838 & \underline{0.9858} & \textbf{\underline{0.9918}} & 0.9200 & \underline{0.9899} & \underline{0.9866} & 0.9804 & \underline{0.9907} \\
     & $C_{\text{area}}\uparrow$ & 0.9842 & 0.9837 & 0.9689 & \underline{0.9888} & 0.9816 & \underline{0.9843} & \textbf{\underline{0.9946}} & 0.9247 & \underline{0.9922} & \underline{0.9883} & \underline{0.9858} & \underline{0.9930} \\
     & Spearman ($\rho$)$\uparrow$ & 0.8077 & 0.9602 & 0.7332 & \underline{0.9629} & 0.8792 & 0.8845 & \textbf{\underline{0.9967}} & 0.8535 & \underline{0.9945} & 0.9297 & 0.9438 & \underline{0.9927} \\
     & Silhouette ($C_{\text{sil}}$)$\uparrow$ & 0.6383 & 0.6056 & 0.3978 & 0.6000 & 0.6165 & 0.6031 & 0.5238 & \textbf{\underline{0.7810}} & 0.5274 & 0.6320 & \underline{0.6454} & 0.5402 \\
     & Stress ($S$)$\downarrow$ & 0.5164 & 0.2205 & 0.5408 & \underline{0.1689} & 0.3985 & 0.4769 & \textbf{\underline{0.0334}} & 0.4730 & \underline{0.0458} & 0.4527 & 0.2870 & \underline{0.0734} \\
     & LOF ($L$)$\downarrow$ & 1.0088 & 1.0093 & 1.0155 & 1.0131 & 1.0106 & 1.0129 & 1.0162 & 1.0146 & 1.0163 & \textbf{\underline{1.0080}} & 1.0100 & 1.0166 \\
    \midrule
    \multirow{6}{*}{\shortstack[l]{\texttt{aging}\\ \citep{sole2020single}}} & $T_{\text{area}}\uparrow$ & 0.9783 & 0.9739 & 0.9710 & 0.9170 & \underline{0.9793} & 0.9709 & 0.9073 & 0.8013 & 0.8127 & 0.9708 & \textbf{\underline{0.9797}} & 0.8521 \\
     & $C_{\text{area}}\uparrow$ & 0.9744 & \textbf{0.9767} & 0.9727 & 0.9477 & 0.9727 & 0.9759 & 0.9482 & 0.9070 & 0.8791 & 0.9684 & 0.9731 & 0.8359 \\
     & Spearman ($\rho$)$\uparrow$ & 0.4830 & 0.4686 & 0.5203 & \underline{0.5464} & 0.4312 & 0.4447 & \textbf{\underline{0.9075}} & 0.0757 & \underline{0.6440} & 0.4038 & 0.4617 & 0.3164 \\
     & Silhouette ($C_{\text{sil}}$)$\uparrow$ & 0.2345 & 0.2084 & 0.0893 & -0.1309 & 0.2120 & 0.1987 & -0.0965 & \textbf{\underline{0.8691}} & -0.1293 & 0.1326 & 0.1683 & -0.3474 \\
     & Stress ($S$)$\downarrow$ & 0.4882 & 0.5079 & 0.4746 & 0.5929 & 0.5503 & 0.5477 & \textbf{\underline{0.1956}} & 0.7022 & \underline{0.3825} & 0.5327 & 0.4904 & 0.4839 \\
     & LOF ($L$)$\downarrow$ & 1.0162 & 1.0118 & 1.0192 & 1.0595 & 1.0157 & 1.0186 & \underline{1.0092} & \textbf{\underline{1.0062}} & \underline{1.0078} & 1.0156 & 1.0403 & 1.0178 \\
    \midrule
    \multirow{6}{*}{\shortstack[l]{\texttt{pbmc3k}\\ \citep{chen2018pbmc}}} & $T_{\text{area}}\uparrow$ & 0.8727 & 0.8613 & 0.8627 & \underline{0.8757} & \underline{0.8794} & 0.8599 & 0.8645 & 0.8659 & 0.8570 & \textbf{\underline{0.8820}} & 0.8680 & 0.8589 \\
     & $C_{\text{area}}\uparrow$ & 0.8978 & 0.8995 & 0.8953 & \textbf{\underline{0.9185}} & 0.8924 & 0.8983 & \underline{0.9056} & 0.8970 & \underline{0.9004} & \underline{0.9034} & 0.8889 & \underline{0.9011} \\
     & Spearman ($\rho$)$\uparrow$ & 0.6812 & 0.6907 & 0.6978 & \underline{0.7092} & 0.6233 & 0.6659 & \textbf{\underline{0.8552}} & 0.5838 & \underline{0.8365} & 0.6632 & 0.6680 & \underline{0.8370} \\
     & Silhouette ($C_{\text{sil}}$)$\uparrow$ & 0.4245 & 0.3707 & 0.5320 & 0.5269 & 0.3311 & \textbf{\underline{0.7090}} & 0.3876 & \underline{0.5520} & 0.2409 & 0.4911 & 0.3585 & 0.2418 \\
     & Stress ($S$)$\downarrow$ & 0.4265 & 0.4089 & 0.4074 & 1.2913 & 1.0298 & 1.3026 & \underline{0.3191} & 0.9763 & \textbf{\underline{0.2901}} & 1.1310 & 0.8945 & \underline{0.3501} \\
     & LOF ($L$)$\downarrow$ & 1.0087 & \textbf{1.0060} & 1.0133 & 1.0165 & 1.0090 & 1.0477 & 1.0135 & 1.0110 & 1.0067 & 1.0112 & 1.0129 & 1.0063 \\
    \midrule
    \multirow{6}{*}{\shortstack[l]{\texttt{kidney}\\ \citep{10.1093/bioinformatics/btab503}}} & $T_{\text{area}}\uparrow$ & 0.9774 & 0.9824 & 0.9730 & \textbf{\underline{0.9843}} & \underline{0.9831} & 0.9817 & 0.9194 & 0.9199 & 0.9269 & \underline{0.9841} & 0.9810 & 0.9273 \\
     & $C_{\text{area}}\uparrow$ & 0.9761 & 0.9857 & 0.9723 & 0.9812 & 0.9794 & 0.9775 & 0.9611 & 0.9637 & 0.9639 & \textbf{\underline{0.9860}} & 0.9807 & 0.9663 \\
     & Spearman ($\rho$)$\uparrow$ & 0.5387 & 0.6399 & 0.5377 & 0.5183 & 0.5230 & 0.5345 & \underline{0.8962} & 0.1990 & \textbf{\underline{0.9467}} & \underline{0.7210} & 0.4457 & \underline{0.9467} \\
     & Silhouette ($C_{\text{sil}}$)$\uparrow$ & 0.0678 & 0.1461 & 0.0404 & \underline{0.2306} & \underline{0.2058} & 0.1064 & -0.2730 & \textbf{\underline{0.9392}} & -0.2111 & 0.1330 & -0.0376 & -0.2500 \\
     & Stress ($S$)$\downarrow$ & 0.4068 & 0.3836 & 0.4037 & 0.4544 & 0.4339 & 0.4465 & \underline{0.2658} & 0.8035 & \underline{0.1930} & \underline{0.3645} & 0.4443 & \textbf{\underline{0.1747}} \\
     & LOF ($L$)$\downarrow$ & 1.0127 & \textbf{1.0112} & 1.0213 & 1.0162 & 1.0155 & 1.0187 & 1.0138 & 1.0272 & 1.0130 & 1.0132 & 1.0219 & 1.0127 \\
    \midrule
     \multirow{6}{*}{\shortstack[l]{\texttt{cortex}\\ \citep{nowakowski2017spatiotemporal}}} & $T_{\text{area}}\uparrow$ & \textbf{0.9729} & 0.9627 & 0.9656 & 0.9562 & 0.9342 & 0.9590 & 0.9192 & 0.8050 & 0.9196 & 0.9442 & 0.9700 & 0.9285 \\
     & $C_{\text{area}}\uparrow$ & 0.9647 & 0.9629 & 0.9661 & \textbf{\underline{0.9710}} & 0.9392 & 0.9520 & 0.9524 & 0.7087 & 0.9528 & 0.9543 & \underline{0.9668} & 0.9620 \\
     & Spearman ($\rho$)$\uparrow$ & 0.6536 & 0.6805 & 0.6894 & \underline{0.7353} & 0.6384 & 0.6154 & \underline{0.9077} & 0.1578 & \underline{0.9077} & 0.6220 & 0.6561 & \textbf{\underline{0.9107}} \\
     & Silhouette ($C_{\text{sil}}$)$\uparrow$ & -0.0154 & -0.0234 & -0.0203 & -0.0413 & -0.0673 & \underline{-0.0097} & -0.1745 & \textbf{\underline{0.5723}} & -0.1730 & -0.0747 & \underline{-0.0008} & -0.1501 \\
     & Stress ($S$)$\downarrow$ & 0.3556 & 0.3667 & 0.3372 & 0.3445 & 0.3993 & 0.4196 & \underline{0.2676} & 0.4918 & \underline{0.2678} & 0.3950 & 0.3633 & \textbf{\underline{0.2587}} \\
     & LOF ($L$)$\downarrow$ & 1.0118 & 1.0110 & 1.0199 & \underline{1.0036} & \underline{1.0037} & 1.0130 & \underline{1.0068} & 1.0150 & \underline{1.0067} & \textbf{\underline{1.0015}} & 1.0133 & 1.0122 \\
    \midrule
    \multirow{6}{*}{\shortstack[l]{\texttt{fashion}\\ \citep{xiao2017fashion}}} & $T_{\text{area}}\uparrow$ & \textbf{0.9771} & 0.9584 & 0.9345 & 0.9675 & 0.9078 & 0.9699 & 0.9250 & 0.9170 & 0.9295 & 0.9758 & 0.9666 & 0.9370 \\
     & $C_{\text{area}}\uparrow$ & 0.9740 & \textbf{0.9810} & 0.9693 & 0.9742 & 0.9446 & 0.9670 & 0.9529 & 0.8211 & 0.9684 & 0.9654 & 0.9687 & 0.9767 \\
     & Spearman ($\rho$)$\uparrow$ & 0.7179 & 0.7552 & 0.8690 & 0.6949 & 0.5093 & 0.5833 & \textbf{\underline{0.9329}} & 0.2825 & \underline{0.9239} & 0.6305 & 0.6089 & \underline{0.8766} \\
     & Silhouette ($C_{\text{sil}}$)$\uparrow$ & 0.1776 & 0.1339 & -0.0074 & 0.1718 & 0.1704 & \underline{0.1973} & -0.0288 & \textbf{\underline{0.8388}} & -0.0059 & \underline{0.1835} & 0.1577 & 0.0384 \\
     & Stress ($S$)$\downarrow$ & 0.3248 & 0.3071 & 0.2730 & 0.3601 & 0.3874 & 0.4774 & 0.4428 & 0.6943 & 0.3131 & 0.3531 & 0.3543 & \textbf{\underline{0.2690}} \\
     & LOF ($L$)$\downarrow$ & 1.0159 & \textbf{1.0060} & 1.0292 & 1.0173 & 1.0180 & 1.0318 & 1.0146 & 1.0361 & 1.0091 & 1.0114 & 1.0235 & 1.0097 \\
    \bottomrule
  \end{tabular}
  }
\end{table*}

\subsection{Comparative Analysis}
In this experiment, we study two questions: \textbf{Q1.}  Can generated algorithms outperform  hand-designed SOTA algorithms?    and  \textbf{Q2.} Can generated algorithms address effectively user-specified structural priorities? 
To investigate this, we first ask AutoManifold   to design separate algorithms  to improve specifically on each of the trustworthiness ($T_{\text{area}}$) , continuity ($C_{\text{area}}$), Spearman  ($\rho$), stress ($S$), silhouette ($C_{\text{sil}}$) and LOF ($L$) scores, with the corresponding designed algorithms referred to as Auto-T, Auto-C, Auto-$\rho$, Auto-S, Auto-Sil, and Auto-L, respectively.
We then instruct AutoManifold    to design algorithms that address a general preference for preserving local  structure, followed by  global  structure, without specifying any particular metrics.
This yields  Auto-local and Auto-global algorithm variants. 
Additionally, we let AutoManifold generate algorithms without specifying any preference, expecting a balanced performance across all   criteria, referred to as     Auto  (w/o).

Table \ref{tab:priority-steering} compares performance of AutoManifold-generated   and   hand-designed algorithms. 
In general, AutoManifold without priority constraint can design algorithms with more balanced and stronger performance than existing baselines.
For priority-driven AutoManifold, the prioritised structural property can steer the synthesised algorithm towards the target metric, while the other properties remain within an acceptable range.
These observations are supported by  scores that  are either better than or  lower, yet still acceptable, relative to  baseline performance  on  respective metrics.

Existing methods t-SNE, UMAP and PaCMAP are designed to address local fidelity.
Their performance on preserving global data structure is controlled via their hyperparameters.
As a result, they can achieve high local scores like  $T_{\text{area}}$  and $C_{\text{area}}$, but less satisfactory global scores, e.g., low or negative $C_{\text{sil}}$ indicating poor cluster separation.
In contrast, taking Auto-Sil as an example, by steering towards better cluster separation,  it generates algorithms that  help  viewers to observe a clearer underlying cluster distribution, while still maintaining  satisfactory local fidelity (mostly over 80$\%$).
In Appendix \ref{app:VisResult}, we demonstrate  visualisation results  for Auto-Sil  and Auto-L.

For further comparison, we  tune  hyperparameters of t-SNE, UMAP and PaCMAP  for individual metrics  and compare AutoManifold against the  tuned baselines optimised for the same objectives and   their  default configuration. 
Results on the PBMC3k dataset are reported in Appendix \ref{sota:tune}.
We observe the following from both Table  \ref{tab:priority-steering} and the additional experiments. 
Optimising local metrics ($T$, $C$, or both) yields only modest  gains  for both AutoManifold and tuned baselines, indicating limited headroom for improving local neighbourhood preservation.
Optimising local metrics affects  global metrics differently  across datasets and methods.  AutoManifold exhibits the largest global changes, followed by PaCMAP, while t-SNE and UMAP  change  the least, likely reflecting their progressively less flexible design spaces.
A consistent trade-off exists between   distance preservation ($\rho$, $S$) and cluster separation  (Sil). Overly enforcing cluster separation distorts the original distance relationships.
When optimising a single global metric ($\rho$, $S$ or Sil), AutoManifold  consistently delivers larger improvements than hyperparameter tuning, often outperforming the strongest baseline. The observed trade-offs   reflect    conflicting objectives  between preserving distance ($\rho$  and  $S$) and enhancing cluster separation (Sil)  rather than  algorithm limitation.
When  optimising   outlier reduction ($L$), tuned t-SNE and UMAP tend to overfit  $L$, substantially degrading other metrics. In contrast, AutoManifold and tuned PaCMAP preserve  strong local scores  with less degradation in  $\rho$, $S$ and Sil.
Under balanced objectives, hyperparameter tuning offers only limited gains over the defaults.
In contrast, AutoManifold consistently maintains  competitive local performance, while   improving one or more global properties (typically $\rho$/$S$ or Sil, and occasionally $L$, depending on the datasets).

\begin{table}[htbp]\centering\small
\caption{Monte Carlo experiments over 5 runs.  }
\label{tab:variance}
\setlength{\tabcolsep}{5pt}
\begin{tabular}{lcccccccccc}
\toprule
  & $T\uparrow$ & $C\uparrow$ & $\rho\uparrow$ & Sil$\uparrow$ &  $S \downarrow$ &  $L \downarrow$ & RNX$\uparrow$ & Shep$\uparrow$ & kNN$\uparrow$ & NMI$\uparrow$ \\
\midrule
Auto-$T$ & 0.874 & 0.896 & 0.639 & 0.428 & 0.446 & 1.011 & 0.281 & 0.460 & 0.967 & 0.709 \\
{\scriptsize $\pm$std} & {\scriptsize .005} & {\scriptsize .010} & {\scriptsize .022} & {\scriptsize .067} & {\scriptsize .016} & {\scriptsize .001} & {\scriptsize .019} & {\scriptsize .019} & {\scriptsize .008} & {\scriptsize .020} \\
\midrule
Auto-$\rho$ & 0.858 & 0.898 & 0.806 & 0.303 & 0.509 & 1.013 & 0.229 & 0.734 & 0.900 & 0.531 \\
{\scriptsize $\pm$std} & {\scriptsize .013} & {\scriptsize .010} & {\scriptsize .104} & {\scriptsize .097} & {\scriptsize .241} & {\scriptsize .008} & {\scriptsize .015} & {\scriptsize .152} & {\scriptsize .028} & {\scriptsize .251} \\
\bottomrule
\end{tabular}
\end{table}

\subsection{Held-out Experiments} 

 Table \ref{tab:priority-steering}   evaluates various versions of AutoManifold, each optimised on a single metric or a subset of metrics,  where metrics not used during the design serve as held-out metrics.
In addition,  we let AutoManifold design  algorithms using  PBMC3k  and evaluate  the designed algorithms on the held-out datasets of Kidney  and Cortex.
The design is optimised on an individual metric  ($\rho$   or  Sil)  and then evaluated using a richer set of held-out metrics.
Results are reported in Table \ref{tab:heldout1} along with its analysis in Appendix \ref{app:heldout}. 
Consistent behaviour across held-out datasets and metrics has been observed for AutoManifold.

\subsection{Stability Analysis} 
We examine the stability of AutoManifold by repeating independently each design (driven by a given priority choice) 5 times on   PBMC3k    for individual metrics.
Overall, designs targeting local metrics are highly stable, while those targeting global metrics vary more.
Table \ref{tab:variance} reports  performance variance of Auto-$T$ and Auto-$\rho$  across  different metrics.
Auto-$T$ consistently achieves  low std across all  metrics.
In contrast, Auto-$\rho$  shows noticeably  larger std on global metrics (e.g., $\rho$, $S$,  Shepard,   NMI), while remaining stable on local metrics ($T$, $C$, RNX) and on kNN accuracy.
This difference may reflect the larger and more flexible design space available for optimising global structures than local neighbourhood preservation.
Overall, the results show that AutoManifold offers greater modelling flexibility than hyperparameter tuning of existing embedding algorithms.
The observed trade-offs arise primarily from incompatible objectives rather than limitations of the generated algorithms.
AutoManifold  achieves local preservation comparable to t-SNE and UMAP, while providing flexibility comparable  to, often exceeding, PaCMAP, with  stronger local performance.

\subsection{Ablation Studies}

We conduct  an ablation study  of Auto-$T$ on Iris dataset by removing individual design components, including  embedding difficulty analysis (Agent 2), neighbourhood analysis (Agent 3), affinity design (Agent 4), objective design (Agent 5), hyperparameter refinement (Agent 6), primitive library, and optimisation formulation guidance  in Eqs.  (\ref{embedding_opt_multi})-(\ref{embedding_opt_constrained}).
Table \ref{tab:ablation} reports performance changes across  metrics.
It indicates that difficulty and neighbourhood analyses, and optimisation guidance contribute the most to AutoManifold performance.
In contrast, hyperparameter refinement and primitive library have  mixed effects, suggesting they are less critical.
Notably, the limited effect of primitives seems to be consistent with the non-deterministic behaviour of LLMs, which can generate designs beyond the  provided choices.

Additionally, we conduct system-level ablation on LLM backbone and skeleton contribution and report results in Appendix~\ref{app:backbone_skeleton}, and observe the following:
Removing the whole skeleton degrades the prioritised metric in 12 of 15 dataset-priority pairs.
Single-LLM and skeleton-free multi-agent designs are inconsistent across priorities. 
Replacing Gemini-3.1-Pro by GPT-5.4 or Claude-Opus-4.8 lowers performance on most datasets.

\begin{table}[htbp]\centering\small
\caption{Component-wise ablation with degradation flagged by $-$ for $\uparrow$  and $+$ for $\downarrow$. N denotes insignificant changes and non-degrading results are underlined.}
\label{tab:ablation}
\setlength{\tabcolsep}{4pt}
  \resizebox{\textwidth}{!}{
\begin{tabular}{@{}lcccccccccc@{}}
\toprule
removed & $\Delta T \uparrow$ & $\Delta C \uparrow$ & $\Delta\rho \uparrow$ & $\Delta$Sil$\uparrow$ & $\Delta S \downarrow  $  & $\Delta L  \downarrow$  & $\Delta$RNX$\uparrow$ & $\Delta$Shep$\uparrow$ & $\Delta$kNN$\uparrow$ & $\Delta$NMI$\uparrow$ \\
\midrule
full   & 0.988 & 0.987 & 0.968 & 0.652 & 0.308 & 1.008 & 0.700 & 0.906 & 0.973 & 0.806 \\
\midrule
Difficulty &  -0.007  & -0.008 & -0.140 & -0.064 & +0.007 & +0.004 & -0.013 & -0.093 & -0.007 & \underline{+0.079} \\
Neighbourhood &  -0.012  & -0.010 & -0.023 & -0.088 & +0.032 & +0.006 & -0.109 & -0.022 & N & -0.101 \\
Affinity &  -0.003  & -0.002 & -0.003 & -0.074 & \underline{-0.160} & +0.002 & \underline{+0.033} & \underline{+0.056} & N & N \\
Objective &  -0.004  & -0.002 & -0.003 & -0.097 & \underline{-0.166} & +0.007 & \underline{+0.042} & \underline{+0.059} & N & N \\
Hyperparam-opt & N & \underline{+0.002} & \underline{+0.008} & -0.048 & \underline{-0.183} & \underline{-0.001} & -0.015 & \underline{+0.070} & N & -0.012 \\
Primitive-lib & N & \underline{+0.001} & \underline{+0.004} &  -0.016  & +0.012 & \underline{-0.001} & \underline{+0.011} & -0.007 & N & -0.020 \\
Opt-Guide &  -0.018  & -0.019 & -0.249 & -0.158 & +0.091 & +0.002 & -0.050 & -0.228 & -0.013 & N \\
\bottomrule
\end{tabular}
}
\end{table}

We also   experiment  with two alternative ways to synthesise algorithms using LLMs in Appendix \ref{app:prompt_ablation}.
One is to directly prompt  a single LLM,  while the other is a  skeleton-free
multi-agent system   that does not rely on any embedding design framework. 
They both have more design freedom, but less agentic structure.
 Results show that  these two alternatives are much less effective and less robust for non-target metrics in  priority-conditioned embedding algorithm design.

\section{Conclusion and Discussion}
We have introduced AutoManifold, an LLM-orchestrated multi-agent system for designing  manifold embedding algorithms tailored to user-specified structural priorities. 
By (1) defining structured design template and mathematical formulation for core algorithmic components and steps and (2) constructing a  primitive vocabulary of affinity matrix, objective function, and optimisation solver, we have transformed LLM agents from unconstrained code generators into reliable, priority-conditioned   designers of   new algorithms with compositional novelty.
Comprehensive evaluations across datasets of varying difficulty   demonstrate  that AutoManifold advances beyond hand-designed algorithms and their traditional hyperparameter tuning.
Results in Table  \ref{tab:priority-steering} show that the synthesised algorithms  outperform strong hand-designed  SOTA baselines on the prioritised properties, while keeping the remaining properties within  acceptable ranges.

Our LLM-driven design pipeline is skeleton-based.
It follows  guidance of the  embedding design framework in Section \ref{sec:framework} to ease creation and search challenges, and  constrains  the design   within a fixed primitive library to ensure numerical stability and mathematical correctness during the synthesis phase.
We intentionally adopt  a structured and  interpretable   design space to guide the agents, as this work represents  an initial exploration  of LLM-driven embedding algorithm design.
By restricting the primitives, we aim to  reduce unproductive exploration, while retaining sufficient flexibility for discovering novel algorithmic variants. 
Nevertheless, we acknowledge that these constraints may limit the overall design freedom and restrict the exploration of potentially promising solutions.

Although the search space contains formulations and   configurations that could recover methods such as t-SNE, UMAP, or  PaCMAP,   it is highly unlikely that the LLM agents would reproduce such algorithms exactly, but  generating similar algorithms with similar skeleton configurations. 
As seen in results, AutoManifold is more effective at prioritising global   properties than local ones. 
One   key reason   is that existing baselines  are difficult to outperform under certain evaluation metrics, particularly the local metrics.
We also observe that  local performance is much more prone to changes in algorithm design. 
Therefore, when working on datasets for which maintaining local properties   does  not compromise much global properties, the generated algorithms often possess lower local scores than strong baselines like t-SNE, UMAP and PaCMAP, while yielding only modest improvement in global scores.
In  future work, we will explore more flexible and adventurous  formulations of algorithmic skeleton to broaden design space, discover new design perspective,  and encourage from-scratch algorithmic novelty, and develop  more effective agentic workflow  to tackle  more creative design tasks of higher challenge. 
Building on the insights gained from  this controlled design of  AutoManifold,  another  important future research direction   is to study the trade-offs between design freedom, algorithm quality, interpretability, and reliability, driven by  more open-ended  design space investigation.

So far, all the generated algorithms by AutoManifold are one-shot with optimised hyperparameters.
During the iterative refinement phase, a worker agent takes the implemented algorithm and tunes its parameters, but it does not make any functional changes within the algorithm.
This prevents AutoManifold   from evolving its algorithmic design in an ad-hoc manner.   
To overcome this limitation,   future work will extend the iterative refinement phase, shifting from hyperparameter refinement to systematic algorithmic revision.
Another promising future direction is to explore iterative user interaction,  allowing the user to  dynamically adjust their priorities  during the design process.

\acks{We thank the anonymous reviewers for their constructive feedback and valuable suggestions, which helped improve the quality and clarity of the paper. This work is supported by the Bicentenary PhD Studentship from Faculty of Science and Engineering, University of Manchester.}

\newpage
\appendix

\section{Embedding Evaluation}
 
\label{app:metrics}

We evaluate embeddings through global, local and outlier lenses.
Global scores assess the preservation of pairwise relationships across the whole dataset and the cluster-level organisation of the embedding.
Local scores assess the fidelity of the neighbourhood around each embedded point.
The outlier score tracks isolated points in the embedding space.
Table~\ref{tab:metric-summary} summarises all scores, where the six design scores are provided to the agents as text and drive algorithm design, while the four held-out scores are never provided to the agents and are used only for the held-out, stability and ablation analyses.
All scores are computed by the deterministic Quantitative Analyst engine (Appendix~\ref{app:engines}).
In the following,  we use $d^{HD}_{ij}=\|\mathbf x_i-\mathbf x_j\|_2$ and $d^{LD}_{ij}=\|\mathbf z_i-\mathbf z_j\|_2$ to denote pairwise distances in the high- and low-dimensional spaces, and $\mathcal N^{HD}_k(i)$ and $\mathcal N^{LD}_k(i)$ to denote the sets of $k$ nearest neighbours of point $i$ in the two spaces.

\begin{table}[!htbp]\centering\small
\caption{Summary of evaluation scores. $\uparrow$/$\downarrow$: higher/lower is better. ``Design'': provided to the agents during design; ``Held-out'': used only for evaluation.}
\label{tab:metric-summary}
\begin{tabular}{llccc}
\toprule
Score & Symbol & Lens & Direction & Role \\
\midrule
Spearman rank correlation & $\rho$ & global (distance) & $\uparrow$ & design \\
Stress-1 & $S$ & global (distance) & $\downarrow$ & design \\
Silhouette & $C_{\text{sil}}$ (Sil) & global (cluster) & $\uparrow$ & design \\
Trustworthiness area & $T_{\text{area}}$ & local & $\uparrow$ & design \\
Continuity area & $C_{\text{area}}$ & local & $\uparrow$ & design \\
Median local outlier factor & $L$ & outlier & $\downarrow$ & design \\
\midrule
Shepard correlation & Shep & global (distance) & $\uparrow$ & held-out \\
Normalised mutual information & NMI & global (cluster) & $\uparrow$ & held-out \\
Area under the $R_{NX}$ curve & RNX & local & $\uparrow$ & held-out \\
$k$NN classification accuracy & kNN & label (local) & $\uparrow$ & held-out \\
\bottomrule
\end{tabular}
\end{table}

\textbf{Global Distance Preservation:}
The Spearman distance score \citep{spearman1961proof} is the Spearman rank correlation between the high- and low-dimensional pairwise distances,
\begin{equation}
 \rho = 1 - \frac{6 \sum_{i < j} \left(r_{ij}^{HD} - r_{ij}^{LD}\right)^2}{N_p(N_p^2 - 1)},
\end{equation}
where $r_{ij}^{HD}$ and $r_{ij}^{LD}$ are the ranks of $d^{HD}_{ij}$ and $d^{LD}_{ij}$ among all $N_p=n(n-1)/2$ pairs.
The stress score \citep{venna2006local} quantifies the distortion of pairwise distances,
\begin{equation}
S =
\sqrt{
\frac{
\sum_{i<j} \left(\tilde d_{ij}^{LD} - \tilde d_{ij}^{HD}\right)^2
}{
\sum_{i<j} \left(\tilde d_{ij}^{HD}\right)^2
}},
\end{equation}
where $\tilde d_{ij}^{HD}$ and $\tilde d_{ij}^{LD}$ are the normalised pairwise distances.
The held-out Shepard score summarises the Shepard diagram \citep{shepard1962analysis}, i.e., the scatter of $d^{LD}_{ij}$ against $d^{HD}_{ij}$ over all pairs, by the Pearson correlation coefficient
\begin{equation}
\text{Shep} = \frac{\sum_{i<j}\left(d^{HD}_{ij}-\bar d^{HD}\right)\left(d^{LD}_{ij}-\bar d^{LD}\right)}{\sqrt{\sum_{i<j}\left(d^{HD}_{ij}-\bar d^{HD}\right)^2}\sqrt{\sum_{i<j}\left(d^{LD}_{ij}-\bar d^{LD}\right)^2}},
\end{equation}
where $\bar d^{HD}$ and $\bar d^{LD}$ are the mean pairwise distances.
Unlike $\rho$, which only rewards a monotone relationship between the two sets of distances, Shep also rewards a linear one.
Higher $\rho$ and Shep and lower $S$ indicate better preservation of global pairwise distances.

\textbf{Cluster Separation:} 
The silhouette score \citep{ROUSSEEUW198753} measures cluster separation in the embedding space with respect to the ground-truth class labels.
For point $i$ in class $\text{clust}_i$, let $d_i = \frac{1}{|\text{clust}_i|-1}\sum_{j\in \text{clust}_i, j\neq i}\|\mathbf{z}_i - \mathbf{z}_j\|_2$ be its mean distance to its own class and $d^n_i = \frac{1}{|\text{clust}^n_i|}\sum_{j\in \text{clust}^n_i}\|\mathbf{z}_i - \mathbf{z}_j\|_2$ its mean distance to the nearest other class $\text{clust}^n_i$.
The score is given as
\begin{equation}
    C_{\text{sil}} = \frac{1}{n}\sum_{i=1}^n \frac{  d^n_i-d_i }{ \max(d^n_i,d_i)} \in [-1, 1].
\end{equation}
It is computed on a random subsample of at most 5,000 points.
Higher values indicate better separation, while negative values indicate strongly overlapping classes.
The held-out NMI score measures the agreement between a clustering of the embedding and the ground-truth labels \citep{strehl2002cluster}.
We partition $\{\mathbf z_i\}_{i=1}^n$ into $K$ clusters by $k$-means, with $K$ equal to the number of classes, and compute
\begin{equation}
\text{NMI} = \frac{2\, I(\mathcal Y;\mathcal C)}{H(\mathcal Y)+H(\mathcal C)} \in [0,1],
\end{equation}
where $\mathcal Y$ and $\mathcal C$ are the label and cluster assignments, $I(\cdot\,;\cdot)$ is their mutual information and $H(\cdot)$ is entropy.
Unlike $C_{\text{sil}}$, NMI does not depend on the geometry of the clusters, only on whether the embedding groups points consistently with their labels.

\textbf{Local Neighbourhood Preservation:}  We employ two typical local fidelity scores, i.e., trustworthiness and continuity.  Trustworthiness   \citep{venna2006local} measures how many of the nearest neighbours in the low-dimensional space are  not neighbours in the original high-dimensional space, computed by
\begin{equation}
T(k) = 1 - \frac{2}{n k (2n - 3k - 1)} 
\sum_{i=1}^{n} \sum_{j \in U_i^k} \left(r_{i,j}^{HD} - k\right),
\end{equation}
where $U_i^k$ is the set of points that are among the $k$-NNs of $i$ in the low-dimensional space but not in the high-dimensional space, and $r_{i,j}^{HD}$ denotes the rank of the point $j$ in the high-dimensional ordering around $i$.
Continuity  \citep{venna2006local} complements the trustworthiness,  measuring how many of the true neighbours in the high-dimensional space remain neighbours in the low-dimensional space. 
It computes  the \textbf{recall} of neighbourhood preservation, given as
\begin{equation}
C(k) = 1 - \frac{2}{n k (2n - 3k - 1)} 
\sum_{i=1}^{n} \sum_{j \in V_i^k} \left(r_{i,j}^{LD} - k\right),
\end{equation}
where $V_i^k$ is the set of points that are among the $k$-nearest neighbours of $i$ in the high-dimensional space but not in the low-dimensional space, and $r_{i,j}^{LD}$ is the rank of $j$ in the low-dimensional neighbourhood of $i$. 
To account for varying neighbourhood sizes, we evaluate both scores on the grid $\mathcal K=\{5, 10, 20, 50, 100, 200\}$, retaining the values $k_1<\dots<k_m$ that satisfy $2k < n-1$ (e.g., $k\leq 50$ for Iris), and summarise each curve by its trapezoidal area over $k$, normalised by the range of retained values:
\begin{equation}
T_{\text{area}} = \frac{1}{k_m-k_1}\sum_{i=1}^{m-1} \frac{T(k_i)+T(k_{i+1})}{2}\,(k_{i+1}-k_i),
\end{equation}
and analogously $C_{\text{area}}$; both lie in $[0,1]$.
As the integration is over $k$ rather than $\log k$, larger neighbourhoods carry more weight: on the full grid, $k\geq 100$ accounts for about 64\% of the area.
The held-out RNX score is based on the co-ranking framework \citep{lee2009quality}.
With $Q_{NX}(K)=\frac{1}{nK}\sum_{i=1}^n|\mathcal N^{HD}_K(i)\cap\mathcal N^{LD}_K(i)|$ the average fraction of preserved $K$ nearest neighbours,
\begin{equation}
R_{NX}(K) = \frac{(n-1)\,Q_{NX}(K)-K}{n-1-K}
\end{equation}
rescales $Q_{NX}$ so that a random embedding scores $0$ and perfect neighbourhood preservation scores $1$.
We report the area under the $R_{NX}(K)$ curve on a log-$K$ scale \citep{lee2015multiscale},
\begin{equation}
\text{RNX} = \frac{\sum_{K=1}^{n-2} R_{NX}(K)/K}{\sum_{K=1}^{n-2} 1/K},
\end{equation}
which summarises all neighbourhood sizes and gives most weight to small neighbourhoods; for $n > 1500$, it is computed on a fixed random subsample of 1,500 points.
Higher $T_{\text{area}}$, $C_{\text{area}}$ and RNX indicate better preservation of the original neighbourhoods.
The held-out kNN score is the accuracy of a $k$-nearest-neighbour classifier ($k=10$) that predicts the class label of each point by majority vote over its neighbours in the embedding, evaluated by leave-one-out; it measures whether local neighbourhoods in the embedding are label-consistent, and higher values are better.

\textbf{Outlier Examination:} We employ the local outlier factor (LOF) \citep{breunig2000lof} to measure  how isolated each  embedded point is compared to its neighbours, denoted by $\text{LOF}(\mathbf{z}_i)$.
Values like  $\text{LOF}(\mathbf{z}_i)\approx 1$  indicate normal points, $\text{LOF}(\mathbf{z}_i)> 1$ suggest potential outlier, whereas  $\text{LOF}(\mathbf{z}_i) \gg 1$ flag strong outliers.
We compute the median of  the  factors of all the embedded points, i.e., $L = \text{median}\left( \{\text{LOF}(\mathbf{z}_i)\}_{i=1}^n\right)$.
A higher value of $L$  indicates the existence of a higher amount of outliers in the embedding space.  

\section{User Interaction} 
\label{app:user}

AutoManifold allows explicit user control on which embedding metrics for the algorithm to improve
over.
It asks the user to select one of the eight pre-defined priority options as listed in Table \ref{tab:priority}.
The corresponding prompt is  passed verbatim to the LLM agents as the design directive, and is separately parsed  into a discrete intent that determines the matching design metrics.
This priority  is selected once at the start of the design process, with no iterative user interaction.
Exploring iterative interactive refinement of the priority during the design process is 
 an interesting   direction for future work.

\begin{table}[htbp]\centering\small
\caption{The eight user-selectable priorities and their corresponding prompts.}
\label{tab:priority}
\resizebox{0.95\textwidth}{!}{
\begin{tabular}{p{3.2cm} p{9.3cm}}
\toprule
\textbf{Option (label)} & \textbf{Prompt passed to the agents} \\
\midrule
Balance local \& global & optimize a balance of Trustworthiness@$k$, Continuity@$k$, Spearman, Silhouette, LOF outlier \%, Stress; no single metric prioritized \\
Prioritise local & optimize Trustworthiness@$k$ and Continuity@$k$ above all else; others secondary \\
Prioritise global & optimize Spearman and Stress above all else; others secondary \\
Maximum Trustworthiness & optimize Trustworthiness@$k$ above all else; others secondary \\
Maximum Continuity & optimize Continuity@$k$ above all else; others secondary \\
Maximum Spearman & optimize Spearman above all else; others secondary \\
Maximum Silhouette & optimize cluster separation (Silhouette) above all else; others secondary \\
Maximum LOF preservation & minimize embedding outlier rate (LOF) above all else; others secondary \\
\bottomrule
\end{tabular}
}
\end{table}

\section{Used Datasets and Difficulty Bracketing}

We evaluate on six datasets from the data catalogue exposed by
\texttt{pipeline/datasets.py},  spanning genomics,
  botany, and  vision sectors. 
Each dataset is assigned to a
difficulty bracket, i.e., \textbf{Easy}, \textbf{Moderate}, or \textbf{Hard}, based on the metrics achieved by the three reference methods PCA, UMAP, and t-SNE on that dataset. 
A summary of dataset information is reported in Table \ref{tab:datasets}.
The bracket assignment is intended to
let us report  where  on the difficulty spectrum LLM-designed gains
concentrate.
For each dataset we report the reference-metric profile achieved by UMAP, and t-SNE under their default settings; the bracket assignment in Table~\ref{tab:datasets} is justified by these reference
profiles. Easy datasets are those where all three baselines achieve trustworthiness greater than 0.95  and silhouette score greater than 0.4. Hard datasets are those where t-SNE achieves silhouette score less than 0.2 or trustworthiness less than  0.85. Although PBMC3K differentiates from its bracket in terms of Trustworthiness score, its marginally high Silhoutte score keeps it in moderate bracket.
\begin{table*}[htbp]
  \caption{Difficulty calibration via reference baselines (UMAP,
  t-SNE). $T(k)$ is trustworthiness at $k{=}50$ on the 2-D embedding;
  ``Sil.''~is the silhouette score on ground-truth labels in the 2-D
  embedding.}
  \label{tab:bracket-calibration}
  \centering
   \resizebox{0.9\textwidth}{!}{
  \begin{tabular}{l l c c c c}
    \toprule
    Dataset & Bracket & UMAP $T(k{=}50)$ & t-SNE $T(k{=}50)$ & UMAP Sil. & t-SNE Sil. \\
    \midrule
    Iris                   & Easy       & 0.981 & 0.985 & 0.654 & 0.640 \\
    \midrule
    Aging Human Skin       & Moderate   & 0.982 & 0.985 & 0.291 & 0.223 \\
    Human Kidney           & Moderate   & 0.991 & 0.993 & 0.333 & 0.176 \\
    PBMC 3k                & Moderate   & 0.859 & 0.872 & 0.567 & 0.440 \\
    \midrule
    Cortex Development     & Hard       & 0.972 & 0.976 & -0.023 & -0.001 \\
    Fashion-MNIST          & Hard       & 0.975 & 0.979 &   0.188 &   0.182 \\
    \bottomrule
  \end{tabular}
  }
\end{table*}
\begin{table*}[htbp]
  \caption{Dataset suite, grouped by difficulty bracket.
  $n$ = samples, $d$ = features, $K$ = classes.}
  \label{tab:datasets}
  \centering
\resizebox{0.72\textwidth}{!}{
  \begin{tabular}{l l r r r l}
    \toprule
    Bracket  & Dataset                        & $n$       & $d$    & $K$ & Domain \\
    \midrule
    Easy     & Iris                           & 150       & 4      & 3   & botany \\
    \midrule
    Moderate     & Aging Human Skin (scRNA-seq)   & 8{,}000   & 50     & 13  & genomics \\
    Moderate     & Human Kidney (scRNA-seq)       & 8{,}000   & 50     & 13  & genomics \\
    Moderate & PBMC 3k scRNA-seq              & 2{,}700   & 50     & 8   & genomics \\
    \midrule
    Hard     & Cortex Development (scRNA-seq) & 4{,}129   & 50     & 47  & genomics \\
    Hard     & Fashion-MNIST (sub.)           & 5{,}000   & 784    & 10  & vision \\
    \bottomrule
  \end{tabular}
  }
\end{table*}

\section{Further Experiments and Results}

\subsection{ Visualisation Results Demonstration}
\label{app:VisResult}

\begin{figure}[htbp]
  \centering
  \includegraphics[width=\textwidth]{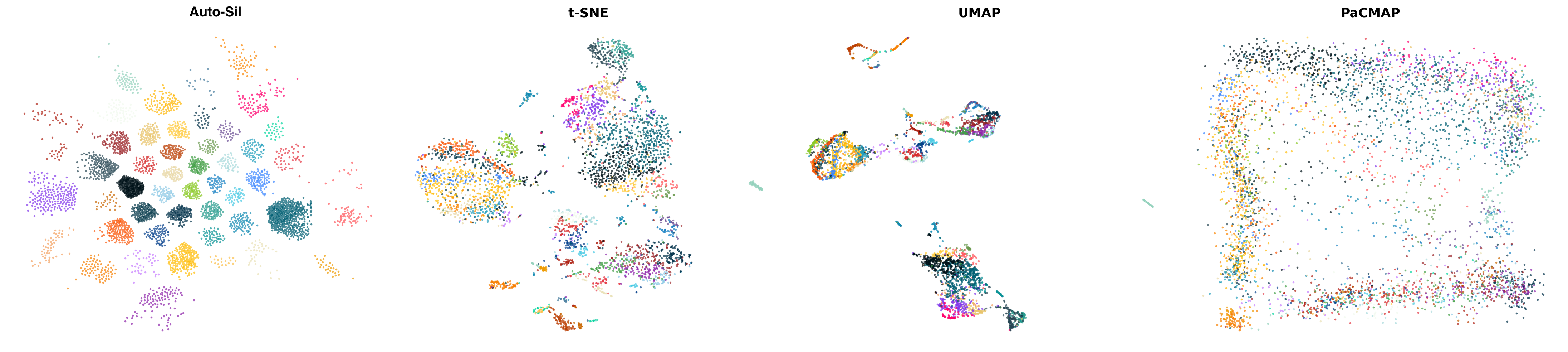}
  \caption{Visualising \texttt{cortex} dataset by the algorithm designed by Auto-Sil, which achieves the highest $C_{\text{sil}}$ (best cluster separation) in Table \ref{tab:priority-steering} among all evaluated methods on this dataset, as well as the three baselines t-SNE, UMAP and PaCMAP at default hyperparameters. }
\label{fig:cortex-auto-sil-vs-baselines}
\end{figure}

Figure \ref{fig:cortex-auto-sil-vs-baselines}  demonstrates  the visualisation results of Auto-Sil for \texttt{cortex} dataset.
Although the   baselines  achieve high local scores with $T_{\text{area}}$ and $C_{\text{area}}$ above 96\%,  they compress the data  into overlapping ribbons with poor cluster separation, e.g., $C_{\text{sil}}$ between $-0.023$ and $-0.015$. 
It is hard to read the underlying cell-type structure from the plot. 
We let Auto-Sil  explicitly steer the algorithm design towards cluster separation.
The agents design a new affinity graph and objective function so that the synthesised algorithm pulls the same cells into compact, visibly disjoint groups, while keeping reasonable local fidelity, e.g., $T_{\text{area}}$ above 80\%.

\begin{figure}[htbp]
  \centering
  \includegraphics[width=\textwidth]{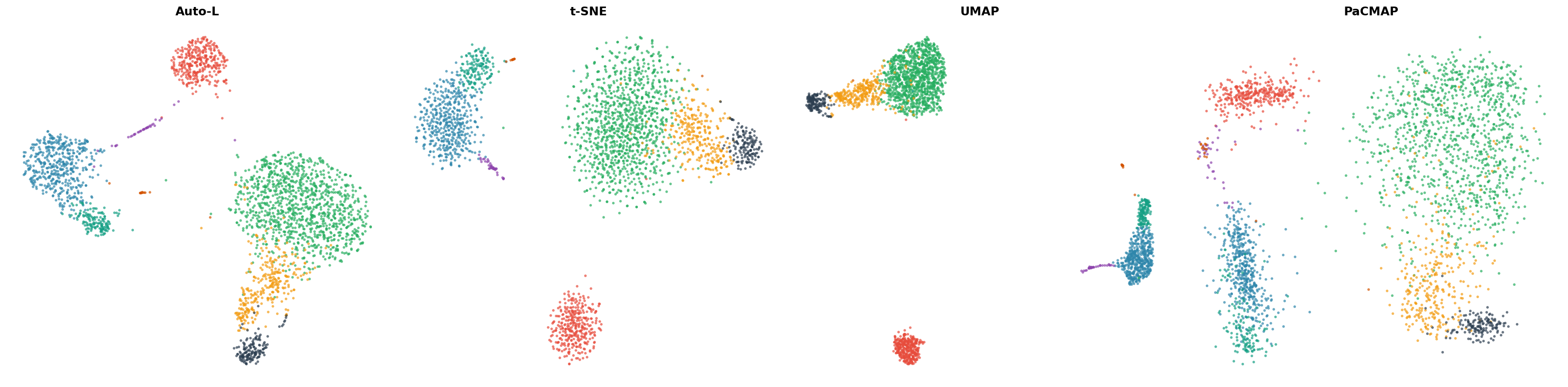}
  \caption{Visualising  \texttt{pbmc3k} dataset by the algorithm  designed by Auto-L, and the three baselines.}
  \label{fig:cortex-auto-l-vs-baselines}
\end{figure}

We demonstrate another visualisation example in Figure \ref{fig:cortex-auto-l-vs-baselines} using \texttt{pbmc3k} dataset,  comparing  baselines with an algorithm generated by Auto-L that aims to reduce outliers. 
It is very interesting to observe that a priority on reducing outliers has led to improvement on local fidelity, indicated by its high underlined $T_{\text{area}}$ and $C_{\text{area}}$ in Table \ref{tab:priority-steering}.
Such phenomenon is also observed for  \texttt{PBMC3K}, \texttt{fashion}, and \texttt{kidney}  datasets as in Auto-L column of Table \ref{tab:priority-steering}. 
It seems that Auto-L might not improve LOF score in all cases despite being conditioned to, but it has the potential to boost local scores. 
Moreover, the visualisation produced by Auto-L in Figure \ref{fig:cortex-auto-l-vs-baselines} reveals  clearer and more highlighted sub-cluster structure, which may facilitate pattern discovery.

\subsection{Comparison with Priority-driven Baseline Tuning}
\label{sota:tune}
AutoManifold  designs algorithms that prioritise one or multiple structural properties each characterised by a score.
We evaluate two  categories of structures: local neighbourhood  preservation  ($T$  and $C$), and global structures measured by   distance preservation ($\rho$  and $S$), cluster separation (Sil), and outlier reduction ($L$).
These objectives are not fully compatible, e.g., improving Sil often degrades $\rho$  and $S$ by distorting  relative distances. 
For a more fine-grained comparison, we   tune  hyperparameters of t-SNE, UMAP and PaCMAP  for individual metrics.
Table \ref{tab:stress} exemplifies results on the PBMC3k dataset, where AutoManifold improves the global metric $S$ without sacrificing  local performance, while   illustrating the inherent trade-off between  distance preservation and cluster separation.

\begin{table}[htbp]\centering 
\caption{Optimising single metric $S$  for  PBMC3k.}
\label{tab:stress}
\resizebox{0.6\textwidth}{!}{
\begin{tabular}{@{}lcccccc@{}}
\toprule
method & $T\uparrow$ & $C\uparrow$ & $\rho\uparrow$ & Sil$\uparrow$ &  $S \downarrow$ &  $L\downarrow$ \\
\midrule
t-SNE (default) & 0.875 & 0.901 & 0.661 & 0.440 & 0.444 & 1.011 \\
\midrule
t-SNE-S   & \textbf{0.857} & 0.879 & 0.654 & 0.318 & 0.385 & 1.007 \\
UMAP-S   & 0.848 & 0.883 & 0.669 & \textbf{0.481} & 0.389 & 1.014 \\
PaCMAP-S   & 0.846 & 0.863 & 0.587 & 0.132 & 0.393 & \textbf{1.005} \\
Auto-S   & \textbf{0.857} & \textbf{0.900} & \textbf{0.837} & 0.241 & \textbf{0.290} & 1.007 \\
\bottomrule
\end{tabular}
}
\end{table}

\subsection{Held-out Experiments}
\label{app:heldout}

We let AutoManifold design algorithms on PBMC3k and evaluate them on the held-out datasets Kidney (K) and Cortex (C).
The designs target a single score, either $\rho$ for distance preservation (Auto-$\rho$) or Sil for cluster separation (Auto-Sil), and are evaluated on all design scores together with the four held-out scores of Appendix~\ref{app:metrics}.
We compare against t-SNE with default hyperparameters and t-SNE tuned on PBMC3k for the same target (t-SNE-$\rho$, t-SNE-Sil).

Table \ref{tab:heldout1} shows consistent behaviour across held-out datasets and metrics.
Auto-$\rho$ achieves strong distance preservation as reflected by $\rho$, $S$ and Shepard, while Auto-Sil achieves competitive cluster separation as reflected by  Sil and NMI.
Auto-Sil improves cluster separation more  on the Kidney dataset (moderate difficulty)  than on  Cortex  (hard), likely because  the design data PBMC3k has a similar level  of difficulty to Kidney (see Table \ref{tab:bracket-calibration} on data information).
The results illustrate  the expected trade-off between preserving global distances ($\rho$, $S$, Shepard) and producing well-separated clusters (Sil, NMI).

\begin{table}[!htbp]\centering
\caption{Held-out experiments on datasets and scores. Algorithms are designed on PBMC3k and evaluated on Kidney (K) and Cortex (C).}
\label{tab:heldout1}
\resizebox{0.9\textwidth}{!}{
\begin{tabular}{l cc cc cc cc cc}
\toprule
   & $T\uparrow$ & RNX $\uparrow$ & $C\uparrow$ &  kNN$\uparrow$ & $\rho\uparrow$ & Shep$\uparrow$ & Sil$\uparrow$ &  NMI$\uparrow$  &  $S\downarrow$ &  $L \downarrow$ \\
\midrule
K, t-SNE  & 0.984 & 0.491 & 0.979 & 0.961 & 0.508 & 0.504 & 0.175 & 0.706 & 0.423 & 1.015 \\
K, t-SNE-$\rho$ & 0.985 & 0.495 & 0.981 & 0.962 & 0.525 & 0.522 & 0.180 & 0.690 & 0.415 & 1.017 \\
K, Auto-$\rho$ & 0.952 & 0.376 & 0.963 & 0.784 & 0.931 & 0.910 & -0.360 & 0.366 & 0.284 & 1.018 \\
\midrule
C,  t-SNE & 0.969 & 0.467 & 0.961 & 0.711 & 0.621 & 0.602 & -0.001 & 0.672 & 0.358 & 1.011 \\
C, t-SNE-$\rho$ & 0.972 & 0.466 & 0.964 & 0.625 & 0.649  & 0.726 & -0.005 & 0.666 & 0.359 & 1.012    \\
C, Auto-$\rho$ & 0.947 & 0.354 & 0.958 & 0.390 & 0.883 & 0.873 & -0.209 & 0.488 & 0.296 & 1.028 \\
\midrule
K, t-SNE   & 0.984 & 0.491 & 0.979 & 0.961 & 0.508 & 0.504 & 0.175 & 0.706 & 0.423 & 1.015 \\
K, t-SNE-Sil & 0.984 & 0.495 & 0.981 & 0.961 & 0.521 & 0.515 & 0.270 & 0.733 & 0.418 & 1.015 \\
K, Auto-Sil & 0.971 & 0.408 & 0.973 & 0.958 & 0.357 & 0.342 & 0.381 & 0.718 & 0.517 & 1.015 \\
\midrule
C, t-SNE & 0.969 & 0.467 & 0.961 & 0.711 & 0.621 & 0.602 & -0.001 & 0.672 & 0.358 & 1.011 \\
C, t-SNE-Sil & 0.964 & 0.459 & 0.951 & 0.705 & 0.605 & 0.565 & 0.002 & 0.664 & 0.367 & 1.011 \\
C, Auto-Sil & 0.962 & 0.388 & 0.958 & 0.657 & 0.663 & 0.589 & -0.007 & 0.653 & 0.479 & 1.009 \\
\bottomrule
\end{tabular}
}
\end{table}

\subsection{LLM Backbone and Skeleton Ablation}
\label{app:backbone_skeleton}

AutoManifold builds on  Google Gemini-3.1-Pro. We report an ablation study on LLM backbone in Table \ref{tab:backbone} by replacing it  with OpenAI GPT-5.4 and Claude Opus-4-8, using the version of AutoManifold that optimises all metrics without hyperparameter refinement. Gemini-3.1-Pro achieves   the strongest and most stable performance across datasets.

To evaluate the contribution of the proposed skeleton, we  remove all  specialist agents and their associated instructions, leaving a blank-canvas multi-agent design system that spawns its self-defined base-agent workers.
Table \ref{tab:skeleton} compares the performance difference (no-skeleton vs. AutoManifold) for algorithms optimised under different metrics on Iris, PBMC3k, and Kidney datasets.
The substantial performance degradation demonstrates the important role of the proposed skeleton in guiding effective design of embedding algorithms.

\begin{table}[!htbp]\centering\small
\caption{LLM backbone ablation (balanced priority, without hyperparameter refinement). \textbf{Bold}: best backbone per dataset and score.}
\label{tab:backbone}
\setlength{\tabcolsep}{5pt}
\begin{tabular}{@{}llcccccc@{}}
\toprule
Dataset & Backbone & $T_{\text{area}}\uparrow$ & $C_{\text{area}}\uparrow$ & $\rho\uparrow$ & Sil$\uparrow$ & $S \downarrow$ & $L\downarrow$ \\
\midrule
\multirow{3}{*}{Iris} & GPT-5.4 & 0.9583 & 0.9690 & \textbf{0.9311} & \textbf{0.4619} & \textbf{0.2166} & \textbf{1.0131} \\
 & Gemini-3.1-Pro & \textbf{0.9620} & \textbf{0.9709} & 0.8239 & 0.4229 & 0.3427 & 1.0139 \\
 & Claude-Opus-4.8 & 0.9261 & 0.8962 & 0.7110 & 0.3483 & 0.4214 & 1.0440 \\
\midrule
\multirow{3}{*}{Aging} & GPT-5.4 & 0.5673 & 0.4992 & $-$0.1919 & $-$0.3797 & 0.9149 & 1.0148 \\
 & Gemini-3.1-Pro & \textbf{0.9661} & \textbf{0.9733} & \textbf{0.4250} & \textbf{0.1455} & \textbf{0.5128} & 1.0179 \\
 & Claude-Opus-4.8 & 0.5293 & 0.5322 & $-$0.0272 & $-$0.6660 & 0.9774 & \textbf{1.0127} \\
\midrule
\multirow{3}{*}{PBMC3k} & GPT-5.4 & 0.5483 & 0.6153 & 0.1961 & $-$0.0422 & 0.6591 & \textbf{1.0154} \\
 & Gemini-3.1-Pro & \textbf{0.8634} & \textbf{0.8809} & \textbf{0.5449} & \textbf{0.4922} & \textbf{0.4372} & 1.0235 \\
 & Claude-Opus-4.8 & 0.7287 & 0.7029 & 0.4206 & $-$0.0836 & 0.7514 & 1.0391 \\
\midrule
\multirow{3}{*}{Kidney} & GPT-5.4 & 0.5040 & 0.5078 & 0.0317 & $-$0.3740 & 0.9418 & 1.0186 \\
 & Gemini-3.1-Pro & \textbf{0.9507} & \textbf{0.9792} & \textbf{0.3992} & \textbf{0.0589} & \textbf{0.4658} & 1.0262 \\
 & Claude-Opus-4.8 & 0.5470 & 0.5684 & 0.0292 & $-$0.7829 & 0.9504 & \textbf{1.0121} \\
\midrule
\multirow{3}{*}{Cortex} & GPT-5.4 & 0.5421 & 0.5292 & 0.1536 & $-$0.2337 & 0.8076 & 1.0168 \\
 & Gemini-3.1-Pro & \textbf{0.9596} & \textbf{0.9691} & \textbf{0.6721} & \textbf{$-$0.0095} & \textbf{0.3495} & 1.0211 \\
 & Claude-Opus-4.8 & 0.5335 & 0.5398 & 0.1516 & $-$0.5710 & 0.9462 & \textbf{1.0140} \\
\midrule
\multirow{3}{*}{Fashion} & GPT-5.4 & \textbf{0.9658} & 0.9746 & 0.6687 & \textbf{0.1624} & 0.3257 & 1.0272 \\
 & Gemini-3.1-Pro & 0.9549 & \textbf{0.9753} & \textbf{0.7429} & 0.0975 & \textbf{0.3124} & \textbf{1.0145} \\
 & Claude-Opus-4.8 & 0.9123 & 0.9289 & 0.5679 & 0.1331 & 0.3526 & 1.0230 \\
\bottomrule
\end{tabular}
\end{table}

\begin{table}[!htbp]\centering\small
\caption{Skeleton ablation, reported as no-skeleton minus AutoManifold. \textbf{Bold}: performance drop after skeleton removal (negative for $\uparrow$, positive for $\downarrow$). \underline{Underline}: the prioritised score of each column.}
\label{tab:skeleton}
\setlength{\tabcolsep}{4pt}
\begin{tabular}{@{}llcccccc@{}}
\toprule
Dataset & Score & Auto (w/o) & Auto-$T$ & Auto-$C$ & Auto-$\rho$ & Auto-Sil & Auto-$S$ \\
\midrule
\multirow{6}{*}{Iris} & $T_{\text{area}}\uparrow$ & \textbf{$-$0.0568} & \underline{\textbf{$-$0.1201}} & \textbf{$-$0.0072} & \textbf{$-$0.0248} & \textbf{$-$0.3610} & $+$0.0016 \\
 & $C_{\text{area}}\uparrow$ & \textbf{$-$0.0560} & \textbf{$-$0.1168} & \underline{\textbf{$-$0.0021}} & \textbf{$-$0.0245} & \textbf{$-$0.3680} & $+$0.0023 \\
 & $\rho\uparrow$ & \textbf{$-$0.0636} & \textbf{$-$0.0366} & \textbf{$-$0.0950} & \underline{\textbf{$-$0.2143}} & \textbf{$-$0.8029} & $+$0.0020 \\
 & Sil$\uparrow$ & \textbf{$-$0.1036} & \textbf{$-$0.2787} & $+$0.0497 & \textbf{$-$0.0036} & \underline{\textbf{$-$0.8109}} & \textbf{$-$0.0013} \\
 & $S\downarrow$ & \textbf{$+$0.1063} & $-$0.0987 & $-$0.1143 & \textbf{$+$0.3324} & \textbf{$+$0.1876} & \underline{$-$0.0124} \\
 & $L\downarrow$ & $-$0.0141 & \textbf{$+$0.0700} & $-$0.0003 & $-$0.0019 & \textbf{$+$0.0248} & $-$0.0008 \\
\midrule
\multirow{6}{*}{PBMC3k} & $T_{\text{area}}\uparrow$ & \textbf{$-$0.0580} & \underline{\textbf{$-$0.3119}} & \textbf{$-$0.0293} & \textbf{$-$0.0079} & \textbf{$-$0.0255} & \textbf{$-$0.0394} \\
 & $C_{\text{area}}\uparrow$ & \textbf{$-$0.0784} & \textbf{$-$0.3255} & \underline{\textbf{$-$0.0323}} & \textbf{$-$0.0056} & \textbf{$-$0.0032} & \textbf{$-$0.1521} \\
 & $\rho\uparrow$ & \textbf{$-$0.0434} & \textbf{$-$0.4470} & $+$0.0951 & \underline{$+$0.0849} & $+$0.2393 & \textbf{$-$0.4040} \\
 & Sil$\uparrow$ & \textbf{$-$1.3619} & \textbf{$-$0.4538} & $+$0.0154 & \textbf{$-$0.0254} & \underline{\textbf{$-$0.1235}} & $+$0.0392 \\
 & $S\downarrow$ & $-$0.3253 & $-$0.5351 & $-$0.8072 & \textbf{$+$0.0909} & $-$0.5500 & \underline{\textbf{$+$0.1382}} \\
 & $L\downarrow$ & \textbf{$+$0.0041} & \textbf{$+$0.0079} & \textbf{$+$0.0358} & \textbf{$+$0.0301} & \textbf{$+$0.0090} & \textbf{$+$0.0067} \\
\midrule
\multirow{6}{*}{Kidney} & $T_{\text{area}}\uparrow$ & \textbf{$-$0.0007} & \underline{\textbf{$-$0.0121}} & \textbf{$-$0.0364} & \textbf{$-$0.0342} & $+$0.0636 & \textbf{$-$0.0068} \\
 & $C_{\text{area}}\uparrow$ & \textbf{$-$0.0023} & $+$0.0047 & \underline{\textbf{$-$0.0093}} & \textbf{$-$0.0358} & $+$0.0153 & \textbf{$-$0.0260} \\
 & $\rho\uparrow$ & \textbf{$-$0.0107} & $+$0.1084 & $+$0.2962 & \underline{\textbf{$-$0.0011}} & $+$0.3135 & \textbf{$-$0.0037} \\
 & Sil$\uparrow$ & \textbf{$-$0.0552} & \textbf{$-$0.3019} & \textbf{$-$0.4189} & \textbf{$-$0.0239} & \underline{\textbf{$-$0.7665}} & $+$0.0092 \\
 & $S\downarrow$ & $-$0.0318 & $-$0.0728 & $-$0.1677 & $-$0.0175 & $-$0.3832 & \underline{$-$0.0104} \\
 & $L\downarrow$ & $-$0.0012 & \textbf{$+$0.0029} & $-$0.0038 & $-$0.0002 & $-$0.0121 & $-$0.0027 \\
\bottomrule
\end{tabular}
\end{table}

\subsection{Alternative LLM Usage Ablation }
\label{app:prompt_ablation}
We experiment with two alternative ways to synthesise algorithms using LLMs: One is to  directly prompt   a single LLM (SingleLLM), while the other is a  skeleton-free multi-agent system (NoSkelAgents) that does not rely on any embedding design framework.
As compared to AutoManifold, they both have more design freedom, but less agentic structure.
\textbf{Single LLM:} The most intuitive while naive way for synthesise embedding algorithm is to prompt a single LLM by a task description written in natural language.
For instance, the LLM is prompted by ``design a 2D manifold embedding algorithm for this dataset and emit the mathematical formulation, pseudo-code, and a self-contained \texttt{custom\_embedding()} Python function'', together with high-level metadata about the dataset (number of samples, number of features, optional class count). 
Under this setting,  there is   no upstream data characterisation,  no   multi-agent decomposition, no  primitive vocabulary,  no  sub-agent spawning, and no  iterative refinement. 
The LLM emits an algorithm in one shot, populating all three representation layers (Section~\ref{sec:method}). 
The code is then validated, executed in the sandbox, and evaluated by the deterministic engines. 
This setting has    maximum  LLM freedom (no constraints) but minimum agentic structure.

\textbf{Skeleton-free Multi-Agent System:} The system is allowed to invent  freely any embedding algorithm using any mathematics, without following any embedding design framework.
It dynamically spawns new specialised sub-agents to gather analyses before finalising its design. 
The   designer must emit   mathematics,  pseudo-code, and executable code coherently.
In this case, the LLM freedom is \emph{maximum}, while the novelty target is \emph{unconstrained}. 
This is equivalent to giving the LLM a blank canvas with no primitive vocabulary constraints.  
Its planner agent  subclasses the planner agent of AutoManifold, inheriting all code-execution helpers, iterative optimisation, visual forensics, and baseline-comparison machinery. 
Only the design and synthesis phases are overridden, ensuring fair comparison across methodologies under
identical evaluation infrastructure.
It implements a unique multi-turn protocol:
\begin{enumerate}
  \item Up to three rounds of LLM interaction.
  \item Per round, the LLM may take one of two actions:
  (a)~\texttt{spawn\_agents}: specifying \texttt{role},
  \texttt{system\_message}, and \texttt{task\_prompt} for one or more
  sub-agents, which the planner instantiates as
  \texttt{DynamicAgent} instances, executes, and whose results are
  appended to the design context; or (b)~\texttt{finalize\_design}: 
  emitting the final algorithm specification in all three layers.
  \item Safety mechanisms: a hard 3-round limit; if no finalisation
  occurs after three rounds, finalisation is forced via a direct LLM
  call; each sub-agent invocation is wrapped in try/except, returning
  \texttt{\{"error": ...\}} as feedback rather than crashing the loop;
  and a fallback minimal ``Adaptive DR'' specification is provided if
  all else fails.
\end{enumerate}
Compared to AutoManifold, it allows three focused code-generation
retries (vs.\ one), allocates five Agents and 10 recovery attempts (vs.\ two),
and falls back to t-SNE if all attempts fail. 
We ask all three LLM-based methods to prioritise the global scores $\rho$ and Stress-1 separately, and tune hyperparameters of baselines to optimise $\rho$ and Stress-1 one at a time as well.
The obtained results are compared in Table \ref{tab:pbmc3k-full-ablation}.
Both single LLM   and multi-agent LLMs without the algorithmic skeleton  do not perform as consistently as AutoManifold.
For instance, SingleLLM-S performs   poorly when optimising $S$, while  NoSkelAgents-$\rho$ is similarly less effective at improving $\rho$.
In contrast, SingleLLM-$\rho$, NoSkelAgents-S and AutoManifold  all achieve strong  performance across most evaluation metrics.

\begin{table*}[t]
  \centering
  \caption{Alternative LLM usage on PBMC3k. Each priority ($\rho$ or $S$) is targeted by SingleLLM, NoSkelAgents and AutoManifold. \textbf{Bold}: best within each priority block.}
  \label{tab:pbmc3k-full-ablation}
  \small
  \setlength{\tabcolsep}{4pt}
  \begin{tabular}{l@{\hskip 6pt}ccc@{\hskip 12pt}ccc}
    \toprule
     & \multicolumn{3}{c}{\textbf{$\rho$-focused}} & \multicolumn{3}{c}{\textbf{$S$-focused}} \\
    \cmidrule(lr){2-4} \cmidrule(lr){5-7}
    Score & SingleLLM-$\rho$ & NoSkelAgents-$\rho$ & Auto-$\rho$ & SingleLLM-$S$ & NoSkelAgents-$S$ & Auto-$S$ \\
    \midrule
    $T_{\text{area}}\uparrow$ & 0.8614 & 0.8433 & \textbf{0.8645} & 0.8075 & \textbf{0.8584} & 0.8570 \\
    $C_{\text{area}}\uparrow$ & \textbf{0.9140} & 0.8786 & 0.9056 & 0.8174 & \textbf{0.9012} & 0.9004 \\
    $\rho\uparrow$ & \textbf{0.8766} & 0.7944 & 0.8552 & 0.6750 & 0.8362 & \textbf{0.8365} \\
    $S\downarrow$ & 0.8175 & 0.4450 & \textbf{0.3191} & 0.5162 & 0.2904 & \textbf{0.2901} \\
    Sil$\uparrow$ & 0.3443 & \textbf{0.4240} & 0.3876 & 0.1762 & \textbf{0.2414} & 0.2409 \\
    $L\downarrow$ & 1.0176 & 1.0350 & \textbf{1.0135} & 1.0249 & \textbf{1.0067} & \textbf{1.0067} \\
    \bottomrule
  \end{tabular}
\end{table*}

\subsection{Difficulty Labelling Sensitivity}
\label{app:difficulty}

Difficulty is assigned per patch rather than per point, since embedding difficulty arises from the relative distance relationships within a local neighbourhood rather than from individual points.
A patch containing outliers receives a higher difficulty level only when the baseline scores indicate that the outliers substantially increase the embedding challenge.
The number of patches is $k=n/m$ for a fixed patch size $m\approx100$, so $k$ scales with $n$; each patch is labelled Easy, Moderate or Hard by the rank of its aggregated reliability score at the 33rd and 67th percentiles.
The labels act as a coarse prior that biases primitive and weight selection, and are subsequently refined by the evaluation and iterative-optimisation stages.
Several generated algorithms also develop their own point-wise difficulty estimate, e.g., by thresholding the distance to the $k$-th nearest neighbour as in DAS-SNE (Appendix~\ref{app:demo}).

Table~\ref{tab:patch-sensitivity} compares the labelling at $m\in\{50,200,300\}$ with the default $m=100$ on four datasets.
Only 35--70\% of points keep their exact three-level label, but this movement is almost entirely reshuffling through the Moderate bucket at the fixed percentile cutoffs.
For the decision-relevant distinction between Easy and Hard, 82--99\% of points (typically above 90\%) keep their classification across all patch sizes.
Patch granularity therefore affects which points lie in the middle band, but not which regions are clearly easy or clearly hard to embed.

\begin{table}[!htbp]\centering\small
\caption{Agreement of difficulty labels with the default patch size $m=100$.}
\label{tab:patch-sensitivity}
\begin{tabular}{@{}lcccccc@{}}
\toprule
 & \multicolumn{3}{c}{Easy--Hard agreement} & \multicolumn{3}{c}{Exact $\{$E, M, H$\}$ agreement} \\
\cmidrule(lr){2-4}\cmidrule(lr){5-7}
Dataset & $m{=}50$ & $m{=}200$ & $m{=}300$ & $m{=}50$ & $m{=}200$ & $m{=}300$ \\
\midrule
PBMC3k  & 0.952 & 0.987 & 0.978 & 0.45 & 0.70 & 0.53 \\
Kidney  & 0.862 & 0.896 & 0.817 & 0.38 & 0.38 & 0.35 \\
Cortex  & 0.914 & 0.958 & 0.952 & 0.46 & 0.52 & 0.61 \\
Fashion & 0.900 & 0.983 & 0.940 & 0.44 & 0.58 & 0.50 \\
\bottomrule
\end{tabular}
\end{table}

\section{Further Implementation Details}
\label{app:impl}
\sloppy

\subsection{Primitive Vocabulary} 
We define primitive vocabulary to communicate model configuration options to   agents, including  six  primitives $\{\text{OptA}_i\}_{i=1}^6$ for constructing  affinity matrix, five primitives $\{\text{OptC}_i\}_{i=1}^5$ for specifying   objective function,  and four primitives $\{\text{OptS}_i\}_{i=1}^4$ for optimisation solvers. 
These are all defined through natural language. 
Specifically, the affinity primitives  include
  \texttt{euclidean\_knn} (standard $k$-NN graph; easy regions),
  \texttt{mutual\_knn} (symmetric $k$-NN; noise-robust),
  \texttt{snn} (shared nearest neighbour; hard or noisy regions),
  \texttt{geodesic} (graph shortest-path distances; curved manifolds),
  \texttt{lof\_weighted} (LOF-down\-weighted edges; outlier-heavy
  regions), and \texttt{learnable} (a neural network metric for complex non-Euclidean structure).
The objective primitives  include
  \texttt{kl\_divergence} ($\mathrm{KL}(P \| Q)$ as in t-SNE; local),
  \texttt{cross\_entropy} (binary cross-entropy as in UMAP; local with
  implicit negative sampling), \texttt{spectral} (Laplacian trace;
  global), \texttt{force\_directed} (spring-electrical model; visual
  interpretability), and \texttt{mutual\_information} (statistical
  dependence).
The optimisation solver primitives  include
  \texttt{sgd}, \texttt{adam}, \texttt{spectral\_eigen}
  (eigendecomposition for spectral costs), and \texttt{fft\_accelerated}
  (FFT-approximated gradients for large-scale data).
We use \texttt{hybrid affinity} and \texttt{multi-scale cost} to distinguish   constant weights from  weight functions (indicated by \texttt{adaptive weighting}), in affinity matrix and objective function construction, respectively.
This together with the other two options of \texttt{constrained multi-objective optimisation} for Eq. (\ref{embedding_opt_constrained}) and \texttt{iterative refinement} for Eq. (\ref{eq:iterative_refine}) constitute the   composition  primitives, controlling both affinity and optimisation formulation.

\subsection{Compute Engines}
\label{app:engines}

A core contribution is the separation of \emph{creative / analytical
tasks} (delegated to LLM agents) from \emph{objective measurement}
(performed by deterministic engines). LLM agents are
(a) \emph{constrained} to a narrow role via targeted system prompts (e.g.\
Agent~7 is purely diagnostic and is explicitly forbidden to propose
parameter changes); (b) \emph{schema-bound}, with required JSON output
structures validated per agent; (c) \emph{grounded}, in that they receive
deterministic metric outputs verbatim and cannot fabricate or manipulate
computed metrics; and (d) \emph{auditable} through mandatory
chain-of-thought fields (e.g.\ Agent~6 must populate the
\texttt{dendrogram\_comparison} and \texttt{visual\_inspection} fields
\emph{before} the \texttt{recommendations} field).

\textbf{Embedding \& Visualisation engine:} \texttt{DRVisualizationEngine}
supports t-SNE (sklearn), UMAP, PaCMAP, and PCA, with an optional
\texttt{n\_pcs} pre-reduction parameter, and renders publication-quality
scatter plots with a 12-colour palette, per-class legends, and the
quality-score badge. It produces deterministic embeddings, serves as
the fallback when LLM-generated code fails, and renders all
visualisation outputs.

\textbf{Quantitative Analyst engine:} \texttt{QuantitativeAnalystEngine} computes global preservation (Spearman~$\rho$ and  stress $S$), local preservation ($T_{\text{area}}$, $C_{\text{area}}$), cluster separation (silhouette, sub-sampled to $5{,}000$ points), outlier statistics (LOF median), and pointwise metrics (per-point trustworthiness, continuity, stress, and a composite distortion weighted as $0.35\,T + 0.35\,C + 0.30\,S$). All values are passed verbatim to agents, so agents cannot fabricate metrics.

\textbf{Clustering engine:} \texttt{ClusteringEngine} constructs
UPGMA dendrograms on per-class centroids (or a sub-sampled raw-data
fallback when no labels are present), represents them as Newick strings
with branch lengths and scipy linkage matrices, and computes
Robinson--Foulds distance, cophenetic correlation, and topology-change
detectors (closest-pair changes, nearest-neighbour inversions,
merge-order disruptions with $\Delta \geq 2$ steps), together with
DR-method-aware calibration notes.

\subsection{Shared Infrastructure }

All worker agents inherit from a common \texttt{BaseAgent} that enforces:
(i) structured JSON-only LLM communication, (ii) multi-strategy JSON
parsing with retry, (iii) coercion of malformed responses, (iv) optional
schema validation hooks, and (v) skill / plugin usage tracking. A unified
\texttt{LLMClient} interface supports OpenAI, Anthropic (Claude), and
Google Gemini providers, with provider-level enforcement of JSON output
(\texttt{response\_format} for OpenAI, assistant-prefill for Anthropic,
\texttt{response\_mime\_type} for Gemini), exponential-backoff retry up to
three attempts, and base64-image vision support. A central
\texttt{PipelineConfig} centralises all hyperparameter defaults, convergence
thresholds (target implicit score $\geq 8.0/10$, patience $= 2$
non-improving iterations, hard cap of $10$ iterations), metric parameters
($k = 10$ neighbours, LOF threshold $1.5$), and per-mode settings.

\subsection{Workflow}
\label{app:workflow}

Each worker is a single-shot, schema-constrained language-model call, a stateless mapping from context to a validated specification, which performs no embedding, no metric computation, and no tool calls of its own. 
Control flow and state are handled by a finite-state Planner (with a language-model backbone for
context management): a deterministic phase graph fixes \emph{what} runs and in \emph{what
order}, and the Planner owns and invokes the deterministic engines, injecting their
outputs verbatim into the next agent's context. 
This enforces grounding: an agent can reason about a metric only because it was measured deterministically and placed in its
context, and cannot recompute or fabricate it. 
The only feedback loops are at the orchestration level: a propose/execute/measure/feed-back refinement cycle, and a bounded
error-recovery loop with baseline fallback. 
This fixed workflow is a deliberate reliability choice.

\subsection{Code Validation and Correction Pipeline}

We introduce a code validation and correction
pipeline operating on executable code, ensuring fair comparison
and addressing the fundamental challenge of LLM-generated code
reliability.
These include the following:
\textbf{Generation Requirements:} Agent~8 must produce a complete
\texttt{def custom\_embedding(data, labels, params)} function that
returns an \texttt{np.ndarray} of shape $(n, 2)$, contains all imports
inside the function body, handles \texttt{labels=None} gracefully, and
uses \texttt{max\_iter} (not the deprecated \texttt{n\_iter}) for
sklearn t-SNE.
\textbf{Automated Pre-execution Correction:} Before execution, the
pipeline applies regex-based fixes for persistent LLM hallucination
patterns, notably the \texttt{n\_iter}~$\to$~\texttt{max\_iter}
substitution.
\textbf{Sandboxed execution.} Code is executed in an isolated
namespace, with a \texttt{ThreadPoolExecutor}-based timeout (60\,s for
small-sample validation, 300\,s for production), and the output is
validated for type, shape, NaN/Inf, and row-count consistency.
\textbf{Small-sample validation.} Before running on the full dataset,
the function is executed on a 500-row stratified subsample with a
60-second timeout. A binary pass/fail decision with a diagnostic
message is produced.
\textbf{Error-recovery loop.} On failure, Agent~10 receives the
failing code, the traceback, the data shape, and the list of previous
attempts; it emits \texttt{diagnosis}, \texttt{fix\_summary},
\texttt{fixed\_code}, \texttt{fixed\_params}, and \texttt{confidence}.
The fix is validated on the small sample; if it fails, the new error is
fed back to Agent~10 (cascading diagnosis). The loop runs up to two
attempts in SingleLLM, Optimize, and Novel modes and up to five in
NoSkelAgents mode. If all attempts fail, the planner falls back to a
deterministic engine: the standard DR engine in Optimize mode, and
t-SNE in SingleLLM, Novel, and NoSkelAgents modes.
\textbf{Runtime code updates during optimisation.} Agent~6 may
propose a \texttt{code\_update} field during the optimisation loop:
the proposed function replaces the current one only after passing
small-sample validation; otherwise the previous code is kept. These updates are limited to hyperparameter related changes, for the cases Agent 6 needs to modify an existing hyperparameter that is not declared in the code. Stale iterations (no parameter or code change) are flagged in history to
prompt bolder changes.

\subsection{Iterative Optimisation Loop}
\label{app:loop}

\paragraph{Loop structure:} For each iteration up to a maximum of ten:
\begin{enumerate}
  \item Compute the embedding (custom code or the standard embedding engine).
  \item Compute deterministic metrics (\texttt{QuantitativeAnalystEngine}).
  \item Compute hierarchical-clustering dendrograms
  (\texttt{ClusteringEngine}).
  \item Render the scatter plot with a quality-score badge.
  \item Agent~7 produces an independent diagnostic score
  (no recommendations).
  \item Agent~6 produces \texttt{implicit\_score},
  \texttt{explicit\_score} (inherited from Agent 7), recommendations, and an optional
  \texttt{code\_update}.
  \item Check convergence; if not met, apply recommendations and
  continue.
\end{enumerate}

\paragraph{Dual scoring:} Agent~6 evaluates two independent assessments:
the \emph{implicit score}~$\in [0, 10]$ is a holistic quality judgement; 
the \emph{explicit score} is a weighted
composite $O = \sum_i w_i O_i$ over individual metric components,
with the weights $w_i$ themselves modulated by the user-supplied
priority $\pi$ (e.g.\ a silhouette-priority run upweights $w_{\text{sil}}$).
The \texttt{explicit\_score} is calculated by Agent 7 and the \texttt{implicit\_score}
is emitted by Agent 6, and convergence is driven by the average of two, ensuring both
qualitative and quantitative satisfaction of the success criteria.

\paragraph{Termination criteria:} The loop terminates when the implicit
score reaches $\geq 9.0$, when patience $= 2$ consecutive non-improving
iterations is observed, or when maximum iterations (set to 5) have elapsed.

\paragraph{Anti-stagnation:} If Agent~6's recommendations produce no
parameter change and no code update, the iteration is marked as stale
in history. Agent~6 is provided with the full optimisation history
including $\Delta$-scores, tried parameter combinations, skipped
recommendations, and stale flags, and its prompt explicitly instructs:
``must try BOLD changes if plateauing, must NOT repeat tried
combinations.''

\subsection{Plugins and Skills System}

The framework includes a Superpowers \citep{superpowers} / PUA \citep{tanweai} plugin system (MIT License) for tracking
agent capabilities. A thread-safe singleton \texttt{SkillTracker} with
run-level lifecycle management auto-maps each agent to a skill category:
data-characterisation and design agents to \texttt{brainstorming},
optimisation, novelty-validation, and error-recovery agents to
\texttt{systematic\_debugging}, and the visualisation evaluator to
\texttt{verification\_before\_completion}. A context-triggered
\textbf{PUA mode} escalates agent effort through pressure-based
prompting. Each skill invocation logs the agent name and ID, timestamp,
duration, parameters, success/failure status, and error messages;
logs are auto-saved to \texttt{outputs/skill\_logs/\{run\_id\}\_skills.json}
on run completion. Skills are tracked via a Python context manager
that automatically measures duration and catches exceptions. In
NoSkelAgents mode, the designer can additionally spawn arbitrary
\texttt{DynamicAgent} instances at runtime, each inheriting
\texttt{BaseAgent} infrastructure and the same skill-tracking machinery.

\subsection{Graceful-Completion Guarantees}

All methodologies are designed to degrade gracefully and return an
output embedding rather than crashing the pipeline. Agent-level: a JSON
parse failure triggers retry parsing followed by an LLM JSON-repair
call, ultimately returning an empty dictionary. Code-level: compilation
failures invoke Agent~10 recovery (2--5 attempts) before falling back to
the deterministic engine. Runtime-level: thread-based timeouts (60\,s /
300\,s) trigger the same recovery. Phase-level (NoSkelAgents): every phase is
wrapped in try/except and continues with degraded state. Design-level
(NoSkelAgents): the designer loop's hard 3-round limit forces finalisation,
falling back to a minimal specification if necessary. Final fallback:
mode-specific deterministic logic is always available. The standard
DR engine for Optimize, and t-SNE for SingleLLM, Novel, and NoSkelAgents. The
result dictionary records \texttt{used\_fallback} and
\texttt{fallback\_reason} so that downstream reporting can distinguish
LLM-designed from fallback embeddings.

\subsection{Output Artefacts}

Each pipeline run produces structured per-agent JSON traces
(\texttt{agent\{N\}\_\{name\}.json}), including the three-layer
algorithm description (LaTeX formulation, pseudo-code, executable
function); per-iteration scatter plots (PNG); the optimisation history
and convergence curve (\texttt{optimization\_history.json},
\texttt{convergence\_curve.png}); five visual-forensics diagnostic plots
and \texttt{visual\_forensics.json}; in SingleLLM, Novel, and NoSkelAgents
modes, a baseline comparison (\texttt{baseline\_comparison.json} plus
side-by-side plots versus t-SNE, UMAP, and PaCMAP); the final
executable algorithm (\texttt{custom\_algorithm\_final.py}); the
skill-invocation audit log
(\texttt{skill\_logs/\{run\_id\}\_skills.json}); and a complete
run summary (\texttt{final\_\{mode\}\_report.json}).

\begin{algorithm}[t]
  \caption{Difficulty-Adaptive Silhouette t-SNE (DAS-SNE)}
  \label{alg:das-sne}
  \small
  \begin{algorithmic}[1]
    \Require Data $X \in \mathbb{R}^{n \times d}$; optional labels; perplexity, $k$, silhouette weight $\lambda$, margin $m$, max iterations $T$
    \Ensure 2-D embedding $Y \in \mathbb{R}^{n \times 2}$
    \State Standardize $X$; if $d > 20$ apply PCA to 20 components; $L_2$-normalize.
    \State Compute the $k$-nearest neighbours and their distances for every point of $X$.
    \State Compute Gaussian affinities $P_{\text{gauss}}$ by binary search to match the target perplexity over the kNN graph.
    \State Build the unweighted kNN adjacency $A$ and the SNN affinities $\mathrm{SNN}=\tfrac{1}{k}\,A A^{\!\top}$, setting the diagonal to $0$. \Comment{eq.\,\ref{eq:das-snn}}
    \State For each $i$ compute the distance to its $k$-th neighbour $d^{(k)}_i$, and let $p_{50},\,p_{85}$ be its $50$th and $85$th percentiles.
    \State Set per-point blend weight
           $w_i \gets \begin{cases} 0.15 & d^{(k)}_i > p_{85}\\ 0.05 & p_{50} < d^{(k)}_i \le p_{85}\\ 0 & \text{otherwise}\end{cases}$
    \State Blend affinities $P_{\text{hyb}} \gets (I-W)\,P_{\text{gauss}} + W\,\mathrm{SNN}$. \Comment{eq.\,\ref{eq:das-blend}}
    \State Symmetrize and normalize: $P \gets \max(P_{\text{hyb}}, P_{\text{hyb}}^{\!\top})$, $P \gets P/\sum_{kl} P_{kl}$. \Comment{eq.\,\ref{eq:das-normP}}
    \State Initialize $Y$ from PCA, rescaled by $10^{-4}/\operatorname{std}(Y_{:,0})$.
    \If{labels are not provided}
        \State Generate pseudo-labels with KMeans ($K=15$).
    \EndIf
    \State Multiply $P$ by $12$ (early exaggeration); initialize Adam moments $m\gets 0,\ v\gets 0$.
    \For{$t = 1, \dots, T$}
        \If{$t = 250$}
            \State Divide $P$ by $12$ (end early exaggeration).
        \EndIf
        \State Compute Student-$t$ affinities $Q_{ij} \gets \dfrac{(1+\|y_i-y_j\|^2)^{-1}}{\sum_{k\neq l}(1+\|y_k-y_l\|^2)^{-1}}$.
        \State Compute KL gradient $\nabla_{\!K\!L} Y$ from eq.\,\ref{eq:das-gkl}.
        \State $\nabla_{\!\text{sil}} Y \gets 0$.
        \If{$t > 250$ \textbf{and} $\lambda > 0$}
            \State Recompute cluster centroids $c_\ell$ from $Y$ and (pseudo-)labels.
            \State For every $i$: $a_i \gets \|y_i - c_{\ell_i}\|$, $b_i \gets \min_{\ell\neq \ell_i}\|y_i - c_\ell\|$, $k^*_i \gets \arg\min$.
            \For{each active point $i$ with $a_i - b_i + m > 0$}
                \State $\nabla_{\!\text{sil}} y_i \gets \dfrac{y_i - c_{\ell_i}}{a_i+\varepsilon} - \dfrac{y_i - c_{k^*_i}}{b_i+\varepsilon}$. \Comment{eq.\,\ref{eq:das-gsil}}
            \EndFor
            \If{$\|\nabla_{\!\text{sil}} Y\| > 10^{-8}$}
                \State Rescale $\nabla_{\!\text{sil}} Y \gets \nabla_{\!\text{sil}} Y \cdot \|\nabla_{\!K\!L} Y\| / \|\nabla_{\!\text{sil}} Y\|$.
            \EndIf
        \EndIf
        \State Total gradient $g \gets \nabla_{\!K\!L} Y + \lambda\,\nabla_{\!\text{sil}} Y$. \Comment{eq.\,\ref{eq:das-gtot}}
        \State Update $Y$ with Adam$(g)$; recenter $Y \gets Y - \overline{Y}$.
    \EndFor
    \State \Return $Y$
  \end{algorithmic}
\end{algorithm}

\subsection{Others}
\label{app:others}

\paragraph{Computational Resources:} All experiments ran on iMac (2017 4.2 GHz Quad-Core Intel Core i7). The total storage required for all datasets is approximately 250MB.

\paragraph{Schema Validation:} Agent~4 outputs are validated for the
presence of \texttt{affinity\_design.primary\_primitive},
\texttt{is\_hybrid}, and \texttt{neighborhood\_specification}; Agent~5
outputs are validated for \texttt{cost\_design.primary\_objective},
\texttt{is\_multiobjective}, \texttt{components}, and
\texttt{cost\_design.formulation}.

\paragraph{Baseline Comparison:} Novel mode automatically runs t-SNE,
UMAP, and PaCMAP as baselines on the same data and emits a side-by-side
embedding comparison with a metrics table where each column highlights
the best-performing method. This is the harness against which novelty
is empirically validated.

\paragraph{Baseline Hyperparameters:}
\emph{Default} runs use scikit-learn / library defaults
(t-SNE: perplexity\,$=$\,30, learning\_rate\,$=$\,200, max\_iter\,$=$\,1000;
UMAP: n\_neighbors\,$=$\,15, min\_dist\,$=$\,0.1;
PaCMAP: n\_neighbors\,$=$\,10, MN\_ratio\,$=$\,0.5, FP\_ratio\,$=$\,2.0)
with \texttt{random\_state}\,$=$\,42.
\emph{Tuned} runs perform a 10-trial seeded random search per method with the
balanced objective $0.2\,T_{\text{area}}+0.2\,C_{\text{area}}+0.3\,\rho+0.3\,(1{-}S)$
over the standard knobs (t-SNE: perplexity, learning\_rate, early\_exaggeration,
max\_iter; UMAP: n\_neighbors, min\_dist, spread, repulsion\_strength, learning\_rate;
PaCMAP: n\_neighbors, MN\_ratio, FP\_ratio, lr).

\paragraph{Cross-layer Consistency:}
The synthesiser agent (Agent~8) is required, in a single response, to
emit all three layers and an explicit cross-layer mapping linking each
mathematical symbol in Layer~1 to a pseudo-code symbol in Layer~2 and a
variable name in Layer~3. The novelty validator (Agent~12) operates on
Layers~1 and~2 (not on the source code), so that classification of
novelty cannot be gamed by superficial code differences.

\section{AutoManifold Output Demonstration}
\label{app:demo}

In general, the outputs of AutoManifold include (1) the mathematical formulation underpinning the generated algorithm, (2) pseudo-code of the generated algorithm, and (3) visual forensics of the output plot.  
In this section, we demonstrate the AutoManifold outputs returned by  running Auto-Sil  on the Cortex Development dataset.
The designed algorithm is named as Difficulty-Adaptive Silhouette t-SNE (DAS-SNE), which improves cluster separation while maintaining reasonable local neighbourhood preservation. 

\begin{figure}[t]
  \centering
  \includegraphics[width=0.7\textwidth]{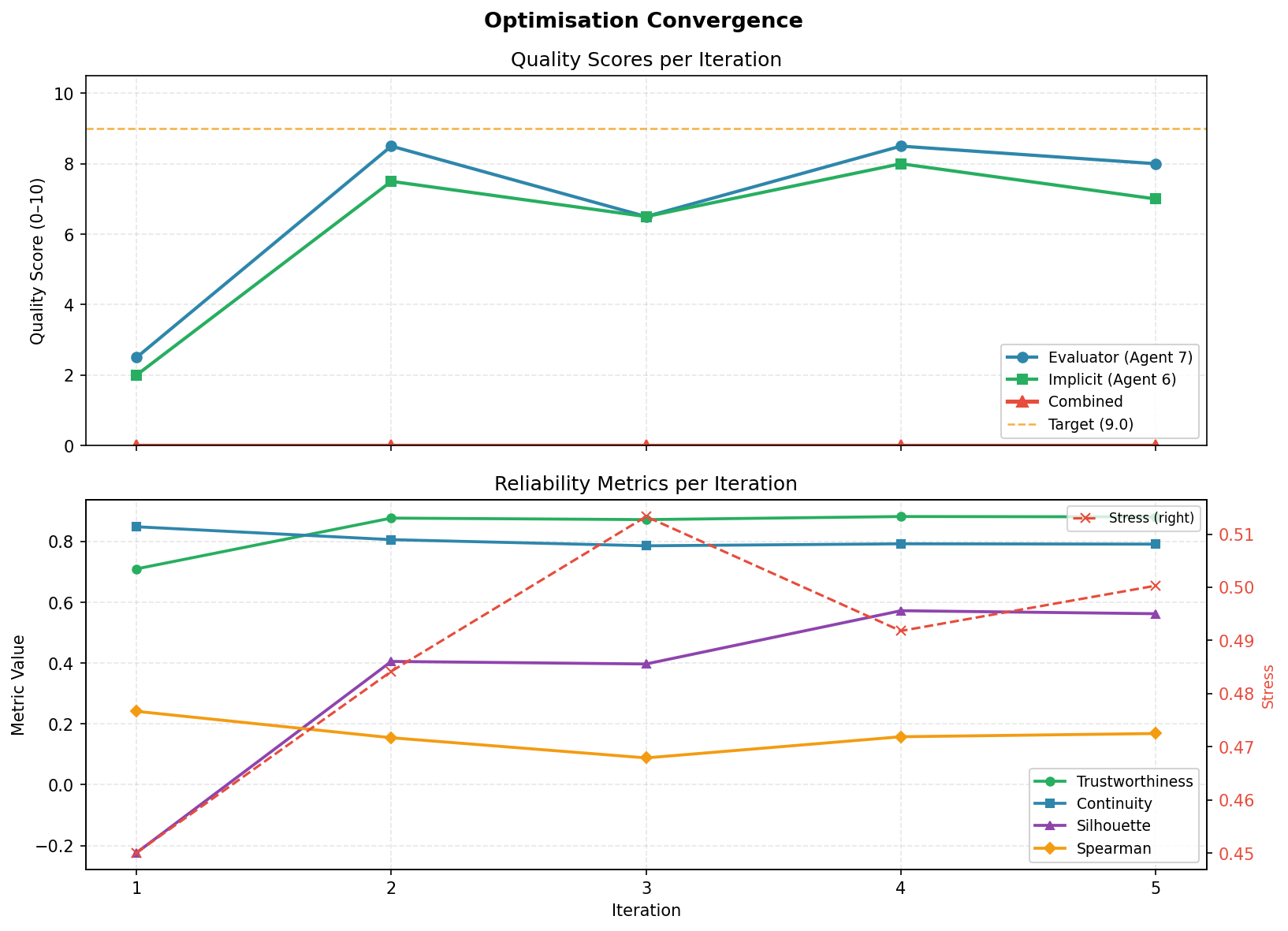}
  \caption{Score changes of Auto-Sil on Cortex Development dataset over 5 iterations of hyperparameter tuning by Agent 6. }
  \label{fig:pair1}
\end{figure}

\paragraph{Mathematical Formulations and Pseudo-code:} We list the  original machine-generated mathematical formulations of DAS-SNE as below.
\begin{align}
  \mathrm{SNN}_{ij} &= \tfrac{1}{k}\,(A A^{\!\top})_{ij},                                            \label{eq:das-snn}\\
  P_{\text{hyb}}    &= (I-W)\,P_{\text{gauss}} + W\,\mathrm{SNN},                                    \label{eq:das-blend}\\
  P_{ij}            &= \frac{\max(P_{\text{hyb},ij},\,P_{\text{hyb},ji})}{\sum_{k,l}\max(P_{\text{hyb},kl},\,P_{\text{hyb},lk})} ,\label{eq:das-normP}\\
  \nabla_{\!K\!L} y_i &= 4\sum_{j}\,(P_{ij}-Q_{ij})\,(1+\|y_i-y_j\|^2)^{-1}\,(y_i-y_j),              \label{eq:das-gkl}\\
  \nabla_{\!\text{sil}} y_i &= \frac{y_i - c_{\ell_i}}{a_i+\varepsilon} \;-\; \frac{y_i - c_{k^*_i}}{b_i+\varepsilon}, \quad\text{if } a_i - b_i + m > 0, \label{eq:das-gsil}\\
  g &= \nabla_{\!K\!L} Y \;+\; \lambda\,\nabla_{\!\text{sil}} Y \cdot \frac{\|\nabla_{\!K\!L} Y\|}{\|\nabla_{\!\text{sil}} Y\|}. \label{eq:das-gtot}
\end{align}
The original machine-generated pseudo-code of DAS-SNE is provided as Algorithm \ref{alg:das-sne}.

\paragraph{Intuitive Explanation:} We  further translate the above mathematical formulations and pseudo-code  into  intuitive explanation of DAS-SNE. 
Starting  from the standard t-SNE  objective $\min_Y KL(P(X), Q(Y))$, DAS-SNE augments $P(X)$ by a second-order neighbouring adjacency matrix $A(X)A^T(X)$. 
It constructs the new $P(X)$ by symmetrising and normalising, given as
\begin{equation}
    \tilde{P}(X) = W(X)\circ P(X) + (1-W(X))\circ A(X)A^T(X), 
\end{equation}
where   $W(X)$ is a data-dependent weighting function. 
Each row of $W(X)$ is assigned a single weight of 0.15 (far), 0.05 (medium) or 0 (close)  determined by embedding difficulty.
Points with distant neighbours are treated as hard to embed, and their neighbourhood connectivity is strengthened through   $A(X)A^T(X)$.
To encourage cluster separation,  DAS-SNE introduces a Silhouette-inspired objective that simultaneously increases within-cluster compactness and  between-cluster separation computed based on cluster centers: 
\begin{equation}
    S(Y) =  \sum_{i=1}^n \left(\frac{\|y_i-c_{i}\|_2^2}{\sum_{k\in I_i}\|y_k-c_{i}\|_2^2 } - \frac{\min_{l\neq clust_i}\|y_i-c_{l}\|_2^2}{\sum_{l=1}^K\|y_i-c_l\|_2^2 }  \right).
\end{equation}
The final optimisation problem becomes
\begin{equation}
    \min_Y KL(\tilde{P}(X), Q(Y)) +\lambda S(Y),
\end{equation}
which is solved using Adam.  
To balance the two objectives, the Silhouette gradient   is normalised by the gradient norm ratio. 
As a result, the Adam updates become 
\begin{equation}
    g= \nabla_{KL} Y + \lambda\frac{\|\nabla_{KL} Y\|_2}{\|\nabla_{S} Y\|_2}\nabla_{S} Y.
\end{equation}
Interestingly, we observe that, in most generated algorithms,  the LLM   does not explicitly formulate optimisation objectives or constraints, but directly designs the gradient update rules together with  optimisation solvers.
The corresponding optimisation objective can often be inferred from the update rules afterwards.

\begin{figure}[t]
  \centering
  \includegraphics[width=\textwidth]{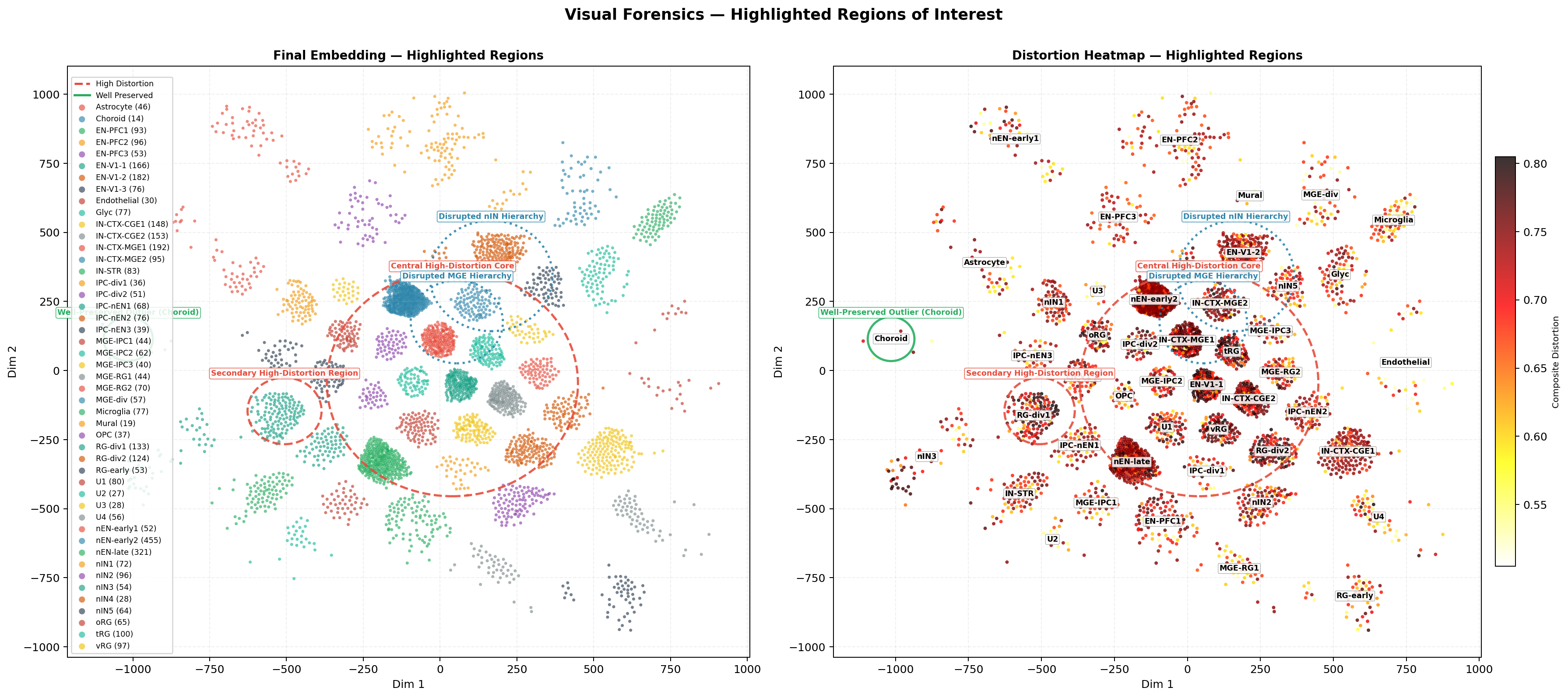}
  \caption{Visual forensics result by  Agent 9.}
  \label{fig:pair2}
\end{figure}

\paragraph{Hyperparameter Search:}  The  hyperparameter setting   recommended by Agent 6 is  provided  in Listing~\ref{lst:das-sne-best-params}.
Figure \ref{fig:pair1} demonstrates the score changes over 5 iterations of  hyperparameter search by Agent 6.
The best result, which is the one retained, is achieved at the fourth iteration in line with the silhouette priority.
The implicit score, i.e., the holistic judgement of Agent 6, and the explicit score, i.e., the priority-weighted metric composite computed by Agent 7 (Appendix~\ref{app:loop}), agree on this choice, indicating that the qualitative assessment of the agents is consistent with the quantitative metrics.

\begin{figure}[htbp]
  \centering
  \begin{lstlisting}[language=json, caption={Agent~6 best-iteration hyperparameters for the synthesised algorithm \emph{Difficulty-Adaptive Silhouette t-SNE (DAS-SNE)} on \texttt{cortex\_development} under the \texttt{max\_silhouette} priority. Selected at iteration 4/5 of the iterative-optimisation loop. Verbatim contents of \texttt{final\_novel\_algorithm\_report.json} $\to$ \texttt{implementation\_result.optimization\_history[best\_iteration].hyperparameters}.}, label={lst:das-sne-best-params}]
{
  "affinity_k":         50,
  "perplexity":         50,
  "learning_rate":      10.0,
  "max_iterations":     2000,
  "silhouette_weight":  50.0,
  "silhouette_margin":  50.0
}
  \end{lstlisting}
\end{figure}

\paragraph{Visual Forensics:} The visual forensics  output  produced by Agent 9 is presented in Figure \ref{fig:pair2}.
It illustrates a distortion heatmap and Agent-9 marked areas, visualising highly distorted, well-preserved and hierarchy-breaking areas on the plot.

\paragraph{Novelty Analysis:}  Agent 12 identifies  DAS-SNE as a novel variant of t-SNE. 
A detailed novelty analysis produced by Agent 12 is illustrated in Listing~\ref{lst:das-sne-novelty}.

\begin{figure}[htbp]
  \centering
  \begin{lstlisting}[language=json, caption={Agent~12 novelty validation output for the synthesised algorithm \emph{Difficulty-Adaptive Silhouette t-SNE (DAS-SNE)} on \texttt{cortex\_development} under the \texttt{max\_silhouette} priority. Verbatim contents of \texttt{agent7\_novelty\_validation.json}.}, label={lst:das-sne-novelty}]
{
  "most_similar_method": "t-SNE",
  "theoretical_differences": "While standard t-SNE relies purely on Gaussian perplexity for high-dimensional affinities and optimizes a strict KL divergence, this algorithm introduces a topology-aware hybrid approach. In the high-dimensional space, it blends the Gaussian probabilities with Shared Nearest Neighbor (SNN) similarities, dynamically increasing the SNN weight in 'hard' or 'medium' difficulty regions to bridge sparse components. In the low-dimensional optimization, it alters the gradient landscape by adding a silhouette surrogate term to the KL divergence, explicitly forcing tighter intra-cluster cohesion and wider inter-cluster separation where the local manifold is difficult to resolve.",
  "suggested_name": "Difficulty-Adaptive Silhouette t-SNE (DAS-SNE)",
  "tagline": "A t-SNE variant that dynamically blends Shared Nearest Neighbors and a silhouette surrogate cost to enhance cluster separation in topologically difficult regions.",
  "is_genuinely_novel": true,
  "recommendations": {
    "next_steps": [
      "Formally define the 'difficulty' heuristic (e.g., local intrinsic dimensionality or density variance) used to classify H, M, and E regions.",
      "Develop a differentiable, computationally efficient approximation for the silhouette surrogate, as exact silhouette scores scale quadratically O(N^2) and are non-smooth.",
      "Analyze the gradient interactions between the KL divergence term and the silhouette surrogate to ensure they do not cause oscillatory behavior during Adam optimization."
    ],
    "potential_limitations": [
      "Calculating Shared Nearest Neighbors (SNN) and the silhouette surrogate significantly increases the computational complexity compared to standard t-SNE.",
      "The adaptive weighting scheme introduces multiple new hyperparameters (thresholds for H/M/E regions) that may be highly dataset-dependent.",
      "Forcing cluster separation via a silhouette surrogate might artificially fracture continuous manifolds or trajectories."
    ],
    "comparison_baselines": [
      "t-SNE",
      "UMAP",
      "PaCMAP",
      "SNN-t-SNE (if available, or a baseline t-SNE with SNN affinities)"
    ]
  }
}
  \end{lstlisting}
\end{figure}

\section{Insight Illustration}

Beyond the reported quantitative scores, AutoManifold includes Agent 9 which performs post-hoc visual forensics by inspecting and analysing the embeddings produced by the generated algorithm (see Figure \ref{fig:pair2}  as an example).
In general, these diagnostics identify which structures are faithfully preserved, where distortion occurs, and which regions of an embedding can be trusted for downstream usage, etc.
We illustrate this analysis for three generated algorithms:  Constrained Adaptive Geodesic Mapping (CAGM)  generated by Auto-$\rho$  on the PBMC3k dataset,   Region-Adaptive Spectral t-SNE  (RAS-t-SNE)  generated by Auto (w/o) under balanced objectives on the Iris dataset,  and   Outlier-Stabilised t-SNE (OS-tSNE) generated by Auto-L on the Kidney dataset.

The insight summary   along with the machine-generated mathematical formulations and pseudo-code summary    for these three algorithms are illustrated below.
\paragraph{RAS-t-SNE:} ``\emph{Formulations:} $
    P=\tfrac{1}{2N}(\tilde P+\tilde P^\top),\; \tilde P=w_{g}P^{\text{gauss}}+w_{geo}P^{\text{geo}},\qquad
    \mathcal{L}=w_{kl}\,\mathrm{KL}(P\|Q)+w_{spec}\,\mathrm{tr}(Y^\top L\,Y),\; L=\mathrm{diag}(P\mathbf 1)-P$.
    Gradient $\nabla_Y\mathcal L=w_{kl}4\!\sum_j(P^{\text{opt}}_{ij}-Q_{ij})(1+\|y_i-y_j\|^2)^{-1}(y_i-y_j)+w_{spec}2LY$;
    complexity $O(N^2)$/iter. \emph{Pseudocode:} (1) perplexity-calibrate $P^{\text{gauss}}$; (2) $k$NN($k{=}15$)$\to$
    shortest-paths $D_{\text{geo}}\to P^{\text{geo}}$; (3) hybrid $\tilde P$, symmetrize, $L$; (4) PCA-init $Y$;
    (5) $t{=}1..T$: early-exaggerate ($4P$, $t{\le}250$), $Q$, KL grad, spectral grad $2LY$, combine, Adam, re-center''.
    On Iris, RAS-t-SNE achieves exact hierarchy preservation, with versicolor and
virginica forming a clade before merging with setosa exactly as in the original
space, so inter-cluster distances in 2D can be interpreted directly. The
diagnostic nevertheless flags residual local distortion concentrated within
setosa, indicating that point-level distances inside that cluster should be
read with more caution than the between-cluster geometry.
    
\paragraph{CAGM:} ``\emph{Formulations:} $D_{\text{target}}=(1-W)D_{\text{geo}}+W D_{\text{snn}}$, $W_{ij}=\tfrac{0.15}{2}(c_i+c_j)$, 
    $G_i  = 2\!\sum_{j\neq i}\!\Big[\alpha\,\mathrm{clip}\!\Big(1-\tfrac{D_{\text{target},ij}}{\|y_i-y_j\|+10^{-4}},-10,1\Big)+\beta^{(t)}\tfrac{P_{ij}}{1+\|y_i-y_j\|^2}\Big](y_i-y_j)$, $
    \beta^{(t+1)} = \beta^{(t)}+\lambda_{lr}\max(0,\,L-D_{\text{target\_knn}})$.
    Complexity $O(N^2\log N+TN^2)$.
     \emph{Pseudocode:} PCA$\to c_i\to$ geodesic+SNN blend $D_{\text{target}}\to$
    affinity $P\to$ Adam descent of $G_i$; every 10 steps raise $\beta$ if local $k$NN spread exceeds target''. 
    On PBMC3k, CAGM recovers a structural diagnostic the baselines do not report:
    it identifies which parts of the embedding preserve the underlying cell hierarchy and which do not. The monocyte lineage is faithfully rendered, with the two monocyte subtypes remaining correctly paired, whereas the T- and B-cell compartments are distorted, since nearby populations are pulled together and
    the merge order among them is inverted. The practical consequence is concrete.
    An analyst can trust sub-clustering within the monocyte lineage in that
    embedding, but should not read fine structure in the T and B compartments as
    biologically meaningful. The baseline embeddings expose no such distinction.

\paragraph{OS-tSNE:} ``\emph{Formulations:} $D_i=\tfrac{s_i-\min s}{\max s-\min s}$ (normalized LOF score),
    $\tilde P=(1-W)\odot P+W\odot L^{\text{lof}},\; W_{ij}=0.15\max(D_i,D_j),\; L^{\text{lof}}_{ij}\propto P_{ij}e^{-|D_i-D_j|},\qquad
    \mathcal{L}=\mathrm{KL}(\tilde P\|Q)+\lambda\sum_i D_i\|y_i\|^2$.
    Gradient $\nabla_{y_i}\mathcal L=4\!\sum_j(\tilde P_{ij}-Q_{ij})(1+\|y_i-y_j\|^2)^{-1}(y_i-y_j)+2\lambda D_i y_i$;
    complexity $O(N^2D+TN^2)$. \emph{Pseudocode:} (1) LOF ($k{=}20$)$\to D_i$; (2) per-point $k_i\in\{15,30,50\}$ by $D_i$
    $\to$ perplexity-calibrated $P$, symmetrize; (3) LOF affinity, blend $\to\tilde P$; (4) PCA-init $Y$;
    (5) $t{=}1..T$: early-exaggerate ($12\tilde P$, $t{\le}250$), $Q$, KL grad, tether grad $2\lambda D_i y_i$, Adam, re-center''.
    On Kidney, OS-tSNE preserves the hierarchy only moderately but achieves strong
local fidelity with very few outlying points. This makes its outlier-stabilising
behaviour explicit and quantified rather than implicit: the analyst learns that
local neighbourhoods are dependable while the global arrangement of clusters is
not.

Across all three cases, the per-point forensic map answers a question the fixed
baselines leave unanswered, namely which regions of this particular embedding
can be trusted, and for what kind of claim.
These insights are not incidental to the algorithms designed; they follow from
what the design pipeline is allowed to see. Baseline methods are optimised
against embedding coordinates alone, so the objectives they can express are
limited to what coordinates encode. Our pipeline instead receives hierarchy
information, the rendered visualisation, and quantitative fidelity metrics
alongside the embedding points. Because the design loop has access to the same
evidence a domain analyst would consult, the algorithms it produces optimise
for, and report on, properties an analyst actually acts upon: which cluster
relationships survive projection, where local structure is trustworthy, and
where it is not. Richer context at design time is what converts a
dimensionality-reduction method into a diagnostic one.

\clearpage
\bibliography{references}

\end{document}